\documentclass[letterpaper,twocolumn,10pt]{article}
\usepackage{zhanggroup}

\usepackage{times}
\usepackage{latexsym}
\usepackage[T1]{fontenc}
\usepackage[utf8]{inputenc}
\usepackage{microtype}
\usepackage{inconsolata}
\usepackage{graphicx}
\usepackage{hyperref}
\usepackage{float}
\usepackage{xspace}
\usepackage{enumitem}
\usepackage{xurl}

\usepackage{booktabs}
\usepackage{natbib}
\usepackage{multirow}
\usepackage{colortbl}
\usepackage{array}
\usepackage{caption}
\usepackage{subcaption}
\usepackage{mdframed}
\usepackage{tcolorbox}
\tcbuselibrary{skins, breakable}

\newcommand{\mypara}[1]{\noindent\textbf{#1.}\xspace}

\begin{document}

\date{}

\title{\bf \texttt{PeerCheck}: Enhancing LLM-Generated Academic Reviews Towards Human-Level Quality}

\author{
Zeyuan Chen\ \ \
Ziqing Yang\ \ \
Yihan Ma\ \ \
Michael Backes\ \ \
Yang Zhang\thanks{Corresponding author.}
\\
\\
\textit{CISPA Helmholtz Center for Information Security} \ \ \ 
}

\maketitle

\begin{abstract}
As academic submissions grow, the traditional peer review process struggles to keep up, raising concerns about quality and fairness.
A trend of using large language models (LLMs) for assistance has emerged.
In this work, we take a critical step toward improving the quality of LLM-generated reviews.
We propose the \texttt{PeerCheck} framework, which investigates LLM-human review differences (\textbf{RQ1}) and explores methods to improve LLM-generated review quality (\textbf{RQ2}).
We first analyzed the human-written reviews with reviews generated by various LLMs and found that LLMs and humans focus on different terms, e.g., LLMs prioritize theory while humans emphasize methodology and experiments.
We further adopt prompt engineering, such as Chain-of-Thought (CoT), and utilize retrieval-augmented generation (RAG) to enhance the LLM-generated reviews towards human-level quality.
We find CoT significantly improves the quality of LLM reviews, while we discover an unexpected ``RAG paradox,'' i.e., experiments with RAG produce different results for various LLMs and, in some cases, even reduce review quality.
Our comprehensive analysis of LLM-generated academic reviews illustrates both possibilities and limitations, contributing to a more effective, human-aligned review system.\footnote{Our dataset is available on \url{https://github.com/TrustAIRLab/PeerCheck}.}
\end{abstract}

\section{Introduction}
\label{intro}

Peer review forms the cornerstone of academic publishing and serves as scientific literature's primary quality control mechanism.
It relies on domain experts who critically evaluate manuscripts for methodological rigor, significance of findings, and clarity of presentation.
Today, academic review faces serious challenges from exponential submission growth and limited reviewer availability, causing delays, inconsistent quality, and reviewer exhaustion~\cite{TDGJWMECPCMCBNRMRKTPFPIIRMMKONKSC17,G11}.

To address this issue, the integration of large language models (LLMs) into manuscript evaluation reflects a growing scholarly trend toward enhancing review processes~\cite{ZZDHL25,DWZDLLZVZSZGLLWLLGXXJWSSGGLWWCRFFLHBCZYY24}.
Advancements in LLMs, particularly models such as GPT~\cite{chatgpt} and Claude~\cite{claude}, have shown remarkable capabilities in understanding, analyzing, and generating text across various domains, including scientific discourse~\cite{ZMHPPCB23,HTK23,IDS23,BBKGPGGLNNH24}.
They exhibit proficiency in comprehending complex concepts and identifying methodological issues~\cite{KGRMI22,ZMHPPCB23}.
With LLM assistance, reviewers can understand manuscripts more efficiently and provide specialized evaluation for emerging interdisciplinary domains~\cite{ZCXJL25,KOPBD25,RJLJKCCHJKKKKKLPYBY25,CYKK26}.
AAAI has even rolled out a new policy that permits LLM assistance in peer review~\cite{aaai_llm}.
Specifically, the LLM takes a structured prompt and the manuscript PDF as input and generates the review according to the instructions.
For example, it can include strengths, weaknesses, detailed comments, and a final decision.
However, such dependence on LLMs would inevitably raise concerns regarding review quality, reliability, and alignment with the expected standards of scholarly evaluation~\citep{SC24,YLMLH25,LRDVW24}.

\begin{figure*}[!t]
\centering
\includegraphics[width=0.90\linewidth]{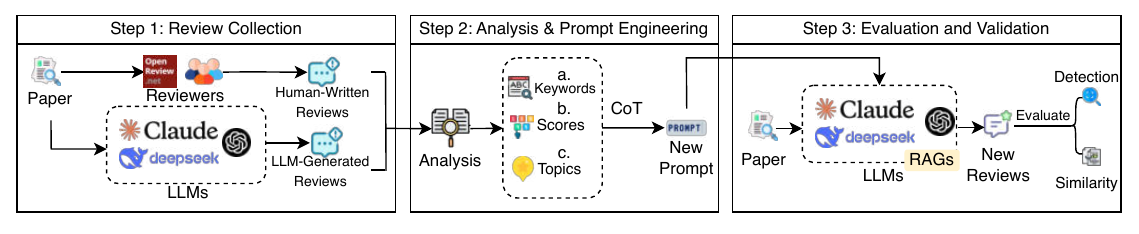}
\caption{Workflow for \texttt{PeerCheck}.
}
\label{figure:workflow}
\end{figure*}

\mypara{Our Contribution}
In this work, we take an important step towards utilizing LLMs for academic peer review.
We formulate two research questions: (\textbf{RQ1}) What are the differences between peer reviews generated by LLMs and those written by humans?
(\textbf{RQ2})
How can we improve LLM-generated review quality to human-level standards?

To address those research questions, we propose a novel framework named \texttt{PeerCheck}.
\underline{Peer} review represents the evaluation of academic work by experts in the same field.
\underline{Check} implies a process of assessment and refinement.
The workflow is displayed on \autoref{figure:workflow}.
Firstly, we collect papers from top conferences as well as their official reviews.
Meanwhile, for each paper, we use three different LLMs with a structured prompt to generate reviews, including GPT-4o, Claude-3.7-Sonnet, and DeepSeek-V3.
Then we analyze and compare human reviews with LLM-generated reviews from the perspective of keywords, rating scores, and topics.
In light of our observations, we conduct prompt engineering and use RAG to enhance the LLM-generated review quality.
We further evaluate the effectiveness of our enhancement across multiple dimensions, including detectability and similarity.
Specifically, detectability assesses how likely LLM-generated reviews can be identified as machine-generated, with lower detectability indicating more human-written reviews.
Similarity serves as one quality indicator, along with human evaluation, as reviews matching the characteristics of effective human feedback are more likely to provide valuable insights to authors.
Results demonstrate that our method can produce more authentic and high-quality academic reviews.
Our contributions can be summarized as follows.
\begin{itemize}[left=0pt]
    \item
    We propose the \texttt{PeerCheck} framework, which
    improves LLM-generated review quality toward human-level standards, increasing GPT-4o's human-like score from 0.379 to 0.645.
    \parskip=0pt
    \item We discover the ``RAG paradox,'' where retrieval augmentation improves GPT-4o but degrades Claude's performance, challenging the ``more information equals better'' assumption.
    \parskip=0pt
    \item We reveal that LLMs exhibit significant ``role sensitivity,'' with Claude-3.7-Sonnet assigning scores of 8 at a rate 57.7\% higher when in the ``PhD student'' role compared to no specified role.
    \parskip=0pt
    \item The RoBERTa-based detector achieves a 0.878 rate for LLM-generated reviews, but our revised Chain-of-Thought (CoT) prompts reduce this by 40.3\%, significantly enhancing LLM-generated review authenticity.
\end{itemize}

\section{Background and Related Works}
\label{background}

\mypara{LLMs in Academic Review}
The peer review faces challenges due to increasing submission volumes, reviewer shortages, and publication delays~\cite{G11,ZCXJL25,KLL25}.
Recent LLM advances offer opportunities to assist or augment parts of the academic review process~\cite{STLKLCHLK25,DWZDLLZVZSZGLLWLLGXXJWSSGGLWWCRFFLHBCZYY24,Y25,RJLJKCCHJKKKKKLPYBY25,LLH24}.
Saad et al.~\cite{SJABIDVB24} explored ChatGPT's potential for peer review assistance, while  Wang et al.~\cite{WBWM24} found substantial overlap between points raised by human reviewers and LLM-generated reviews.
Despite promising outcomes, substantial constraints remain.
 Gao et al.~\cite{GHSa25} found LLMs identify 68\% of methodological issues caught by humans but miss subtle flaws requiring expertise.
Enhancement approaches include Retrieval-Augmented Generation (RAG)~\cite{CLHS24}, Chain-of-Thought prompting (CoT)~\cite{WWSBIXCLZ22}, and hybrid human-AI workflows~\cite{WHJKIH24}.
However, comparisons across multiple advanced models using standardized review criteria remain limited~\cite{SAA25,ZCY24}, and most approaches enhance LLM capabilities rather than targeting academic review weaknesses~\cite{LYXYD25,KOPBD25}.
Our research addresses these limitations through \texttt{PeerCheck}, which enables quantitative comparison and targeted enhancement strategy development.

\mypara{LLM-Generated Content Evaluation and Detection}
As LLMs proliferate, identifying LLM-generated academic reviews has become crucial for maintaining research ecosystem integrity.
Several frameworks assess LLM outputs, including GPTScore~\cite{FNJL25} for general text quality and \citet{CCCLYWGWLLYWXHFZYLCCZZK24}'s multidimensional framework for scientific content.
Their analysis of scientific abstracts showed LLMs often produce scientifically plausible but factually questionable content with fabricated references~\cite{GHSa25}.
Detection grows challenging as LLMs advance, with researchers finding current detection tools ``neither accurate nor reliable''~\cite{WABFGPŠW23}.
More advanced detection methods include multi-feature models analyzing both lexical patterns and semantic structures~\cite{PVÁ24} and watermarking approaches that embed undetectable signals in LLM outputs~\cite{LXGY24}.
However, integrating detection insights with improvement strategies remains underexplored, especially for specialized tasks like academic review.
The characteristics making LLM reviews detectable, like generic criticism patterns and limited expertise signaling, highlight specific areas needing improvement~\cite{YLMLH25,LYXYD25}.
Our research identifies key LLM-human review differences and develops improvements via revised prompts in various LLM settings.

\section{\texttt{PeerCheck}}
\label{method}

We employ \texttt{PeerCheck} (shown in \autoref{figure:workflow}) to evaluate and develop revised prompts for LLMs.

\mypara{Review Collection}
We first collect papers from top conferences and crawl their official peer reviews from OpenReview for comparison with LLM-generated reviews.
Then, we implement a parallel processing architecture utilizing three state-of-the-art LLMs: GPT-4o, Claude-3.7-Sonnet, and DeepSeek-V3.
All three LLMs process authentic research papers from a corpus consisting of 1,920 papers, primarily in the machine learning domains.
Each LLM generates comprehensive reviews of each paper, extracting the strengths, weaknesses, and comments with questions from the paper, and also providing a final rating score for each paper.
The evaluation prompt is shown in Appendix~\ref{evaluation prompt}.
These LLM-generated reviews serve as benchmark outputs for measuring further prompt optimizations.
Following prior work on persona prompting~\cite{ZQCWDHZJ24,SMR23,LN24}, we simulate diverse reviewer personas, including professors, industry experts, and students, to strengthen our analysis framework.
The specific prompts are shown in  Appendix~\ref{roleplay prompt}.
This implementation improves the comprehensiveness and quality of LLM-generated reviews across our measurement framework.
The metrics across all three models offer redundancy and enhance the robustness of information extraction methodologies.

\begin{figure}[!t]
\centering
\includegraphics[width=0.90\linewidth]{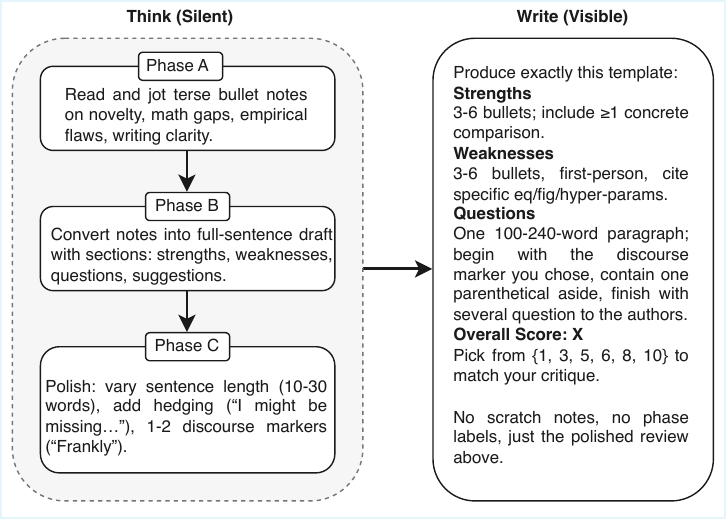}
\caption{
The Chain-of-Thought (CoT) peer review process.
The thinking process contains three reflective phases that help understand and review the manuscripts.
}
\label{figure:cot}
\end{figure}

\mypara{Analysis \& Prompt Engineering}
In our evaluation, prompt engineering plays a crucial role, serving as the foundation for improving the quality and human-likeness of LLM-generated reviews.
In the second step, we employ a systematic approach to develop and revise prompts to bridge the gap between LLM-generated and human-written reviews.
(a) For the keyword analysis, we first extracted distinctive keywords from high-quality human reviews.
The keywords are validated against IEEE VIS criteria~\cite{keywords_selection}, which are established guidelines for assessing visualization research quality.
Then we compared keyword patterns between human and LLM reviews to identify language differences.
(b) Meanwhile, we compared the rating scores given by LLM-generated and human-written reviewers.
This scoring analysis shows how human reviewers and LLMs measure the quality of articles when making acceptance decisions and assigning scores.
(c) For topic analysis, we extract articles receiving high scores ($\ge$ 7) from both human reviewers and LLMs.
We compare topical domains of highly rated papers to analyze topic preference differences between LLM-generated and human-written reviewers, revealing potential disciplinary biases in automated evaluation systems.
After comparing and analyzing, we produce revised prompts directing LLMs to use specific terms and reasoning styles, adopting reviewer personas,
justify scoring criteria, and consider disciplinary context in paper reviewing.
These revised prompts include keywords and examples drawn from human reviews and provide a concrete model for expert peer review in LLMs.
Following the methodological intervention, we employed a CoT~\cite{TMPB23,WWSBIXCLZ22} approach to refine and iterate our prompt formulations, as shown in \autoref{figure:cot}.
This iterative improvement process enables LLMs to mimic human analytical reasoning for generating high-quality reviews.
The revised prompts (Appendix~\ref{enhance prompt}) guide LLMs through a structured evaluation process incorporating linguistic features and reasoning patterns characteristic of human-written reviewers.

\mypara{Evaluation and Validation}
The final step implements a validation module that assesses whether LLM-generated reviews achieve human-level quality.
We measured LLM-generated detectability and similarity to human-written reviews.
In this context, we utilize RAG to enhance LLM review quality by retrieving external knowledge from human reviews, addressing the identified gaps between LLM-generated and human-written evaluations~\cite{CLHS24,LPPPKGKLYRRK20}.
A validation set of 480 research papers, distinct from those used in previous phases, is processed through the enhanced system architecture.
The validation process incorporated two evaluation mechanisms to ensure robust assessment of the revised prompts.
We employ a metric-based detection methodology and a model-based detection that incorporates multiple statistical measures to establish the detection of LLM-generated reviews~\cite{HSCBZ24}.
To see whether LLM-generated reviews by revised prompts can be judged as human-written by detection methods.
For similarity detection, we measure lexical and semantic overlap between LLM-generated reviews and a corpus of the same size as human-written reviews.
Based on these results, we determine whether using revised prompts makes LLM-generated reviews more similar to human-written reviews.
This comprehensive validation approach assessed prompt optimization effectiveness across multiple performance dimensions.

\section{Experimental Setup}
\label{dataset}

\begin{figure}[!t]
\centering
\includegraphics[width=0.90\linewidth]{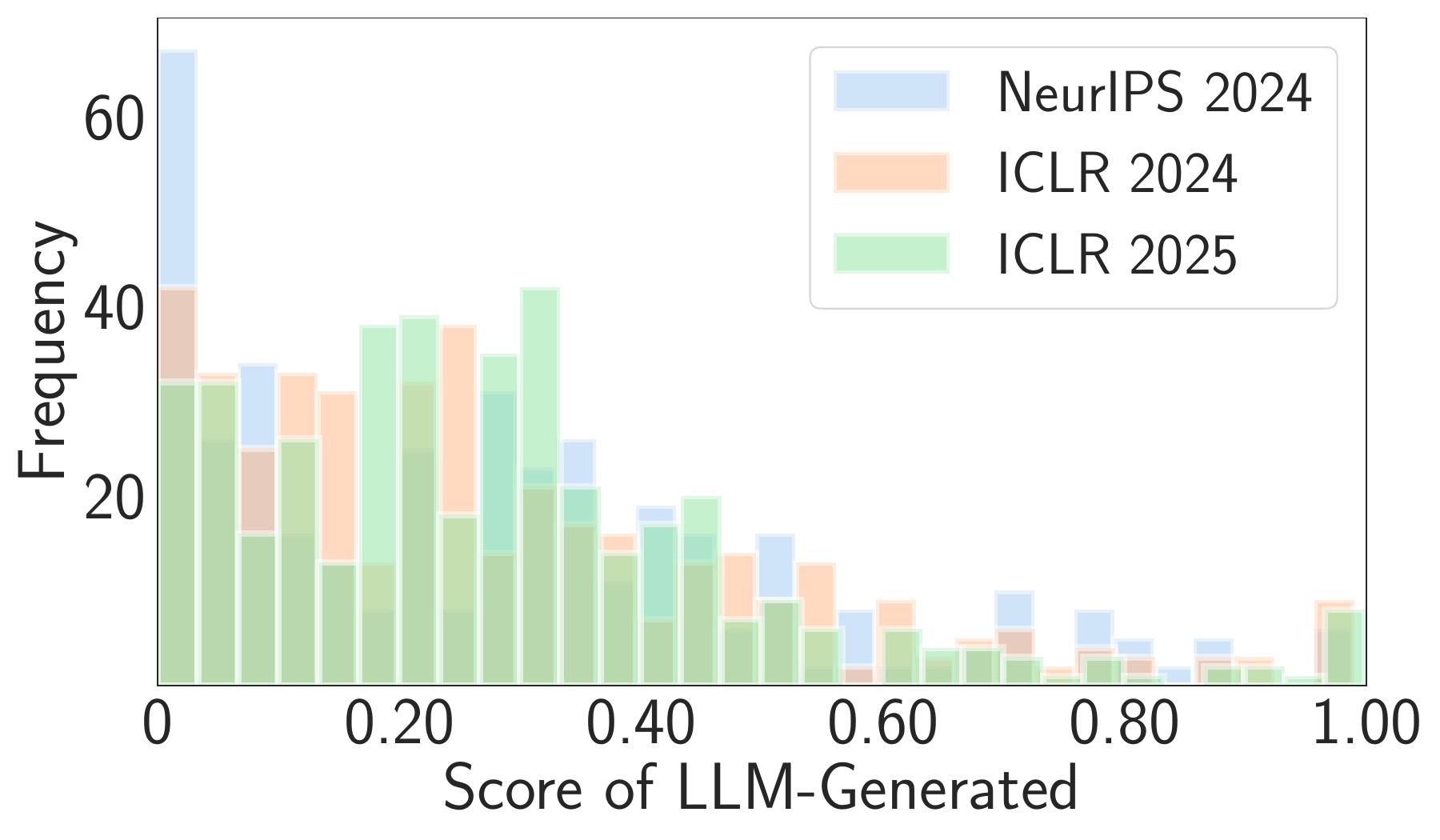}
\caption{
We randomly select 400 reviews from each of the three conferences to detect LLM-generated text.
The histogram shows the frequency distribution of detection scores across NeurIPS 2024, ICLR 2024, and ICLR 2025, with lower RoBERTa-based scores suggesting most reviews are human-authored.
}
\label{figure:human review}
\end{figure}

\mypara{Datasets}
The dataset covers all papers from ICLR 2024/2025~\cite{iclr} and NeurIPS 2024~\cite{neurips}.
We ensure the human-written by filtering them with RoBERTa~\cite{LOGDJCLLZS19},
retaining only those reviews that have more than 80\% probability of being written by humans (\autoref{figure:human review}).
After dividing in an 8:2 ratio, we ultimately formed two datasets of 1,920 papers (analysis) and 480 papers (validation) for comprehensive evaluation across multiple language models.
To evaluate the datasets, we use rating score comparison, semantic similarity comparison, keywords analysis, and topic analysis.
Details are in \autoref{dataset eval}.

\mypara{Evaluated LLMs}
Our study compares the review generation capabilities of three advanced LLMs.
GPT-4o~\cite{gpt4o} is a versatile model known for its strong reasoning abilities in various domains.
Claude 3.7 Sonnet~\cite{claude3.7}, which features a specialized ``thinking mode'' for enhanced analytical reasoning.
DeepSeek V3~\cite{deepseek}, which integrates advanced multimodal capabilities focusing on visual and scientific reasoning.
Each model represents LLM designs and training methods, revealing different academic review generation skills and improvement strategies.

\mypara{LLM Detection and Similarity Metrics}
We use a multidimensional metric technique to evaluate LLM-generated reviews' quality and recognizability by analyzing their detection efficacy and likeness to human-written reviews.
For detection, we use log-likelihood~\cite{HSCBZ24}, log-rank~\cite{SZWN23}, GLTR~\cite{GSR19}, LRR~\cite{SZWN23,HSCBZ24}, and RoBERTa~\cite{LOGDJCLLZS19}.
For similarity, we use BLEU scores~\cite{PRWZ02}, ROUGE scores~\cite{L04}, token-level F1 scores~\cite{DJRLXSW20}, and cosine scores~\cite{LSLW16}.
Details on metrics are provided in \autoref{detection eval}.

\section{Human vs.\ LLM Reviews}
\label{rq1: llm review}

In this section, we compare human-written reviews with LLM-generated reviews across multiple quality dimensions, i.e., to answer \textbf{RQ1}.

\begin{figure}[!t]
\centering
\begin{subfigure}{0.45\columnwidth}
\includegraphics[width=\columnwidth]{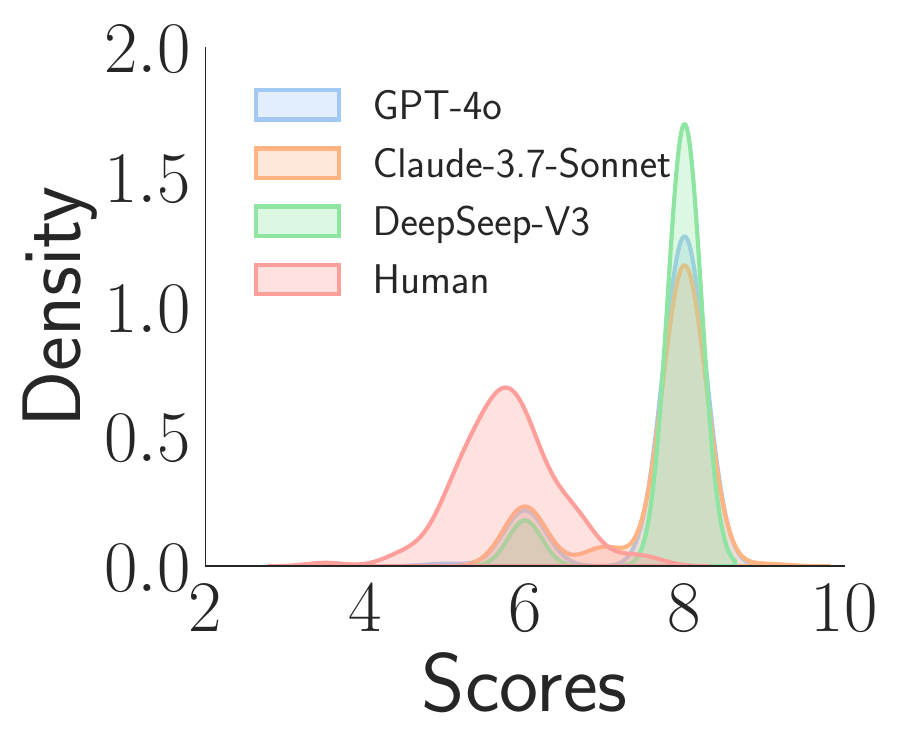}
\caption{NeurIPS Accept}
\label{figure:neurips accept}
\end{subfigure}
\begin{subfigure}{0.45\columnwidth}
\includegraphics[width=\columnwidth]{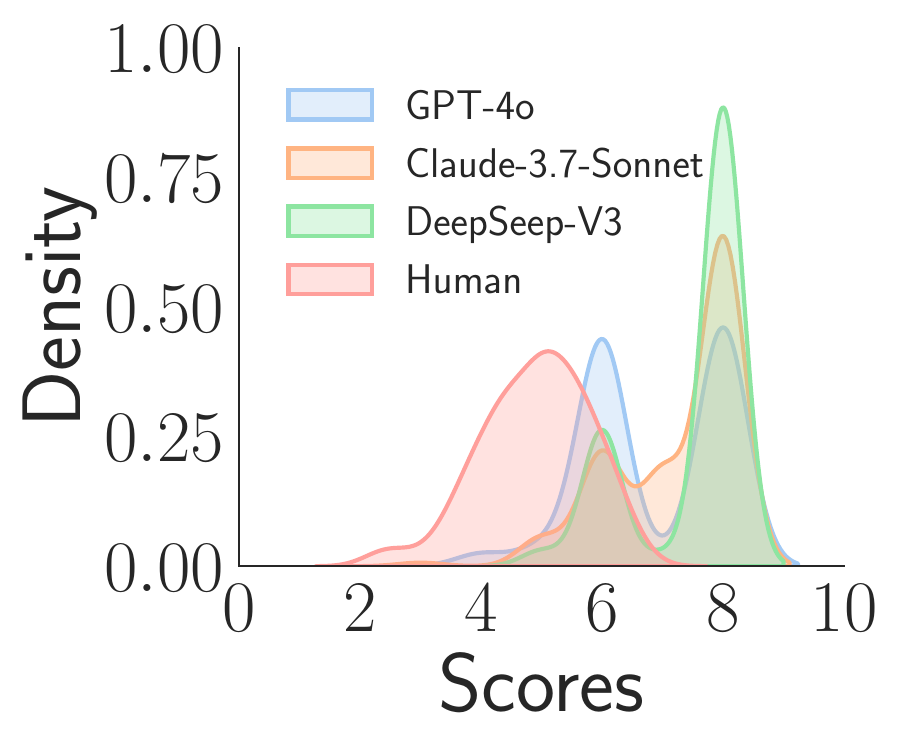}
\caption{NeurIPS Reject}
\label{figure:neurips reject}
\end{subfigure}
\caption{LLMs vs.\ human rating scores in NeurIPS.}
\label{figure:neurips score}
\end{figure}

\mypara{Overall Performance Comparison}
Our comparative analysis of rating scores assigned by LLMs versus human reviewers reveals notable distributional differences, as illustrated in \autoref{figure:neurips score}.
Compare the other two conference rating scores in \autoref{figure:iclr24 score} and \autoref{figure:iclr25 score} of \autoref{additional rq1}.
The human reviewer scoring pattern demonstrates a more concentrated distribution across the evaluation spectrum.
In contrast, the three LLM reviewers exhibit considerably more dispersed scoring patterns, with a pronounced clustering in the 6 to 8 score range.
A striking finding appears: LLMs' judges show broader score distributions than humans, consistently assigning higher scores to rejected papers.
DeepSeek-V3 particularly assigned elevated ratings to both accepted and rejected papers.
Nevertheless, diverse LLMs demonstrate conspicuous disparities in their scoring patterns when evaluating rejected manuscripts.

\begin{figure}[!t]
\centering
\includegraphics[width=0.90\linewidth]{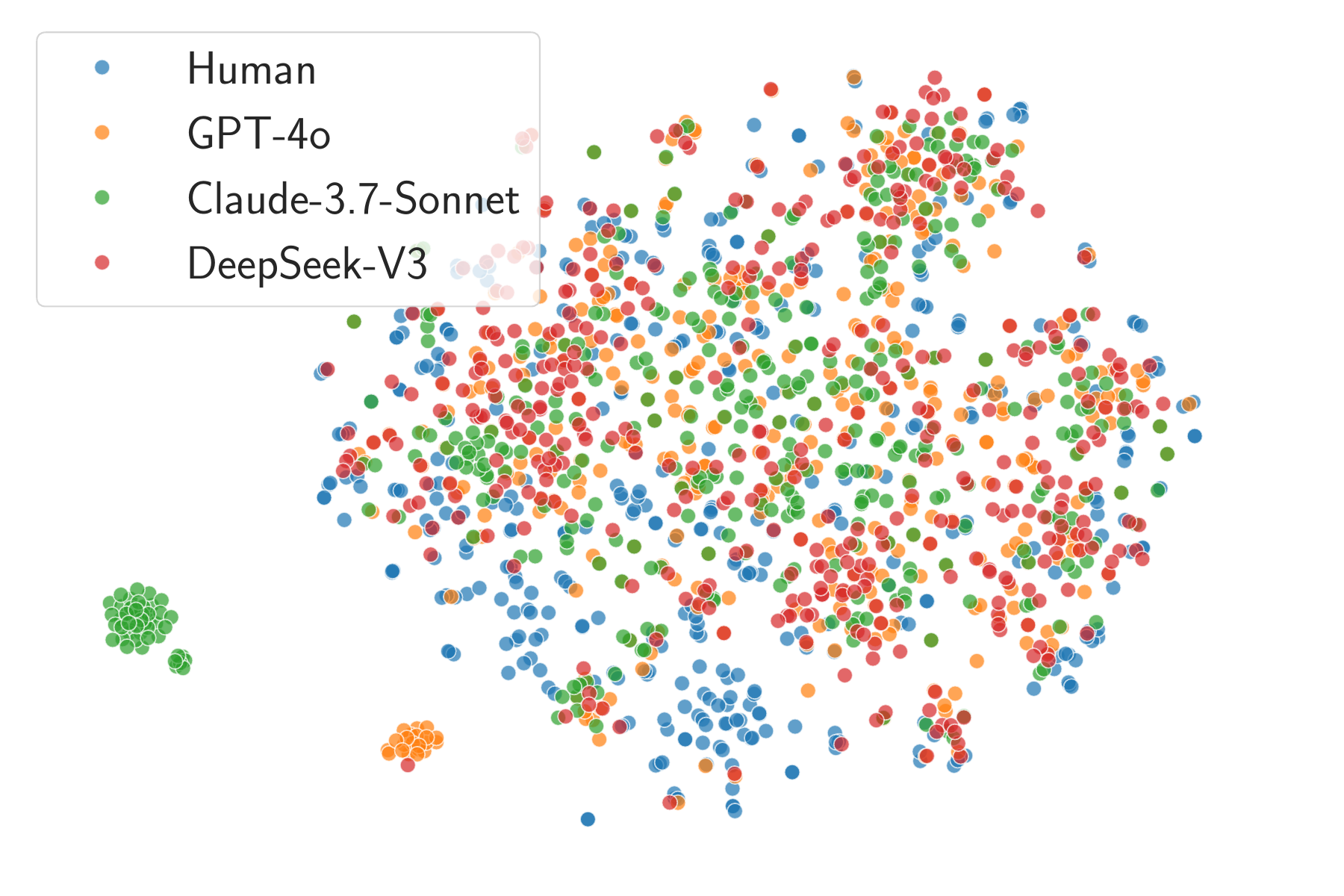}
\caption{
T-SNE visualization showing clustering patterns reflecting linguistic and content differences.
}
\label{figure:tsen}
\end{figure}

For better analysis, we randomly extract a corpus of 400 review comments from each source, i.e., human reviewers and the three LLMs under investigation.
The feature space distribution is visualized through t-SNE~\cite{MH08}, as illustrated in \autoref{figure:tsen}.
The visualization demonstrates extensive mixing of data points across sources, revealing substantial feature space overlap between human- and LLM-generated reviews, thus presenting significant discrimination challenges in most cases.
However, distinct isolated clusters appear in the visualization, particularly from Claude-3.7-Sonnet and GPT-4o reviews, indicating peripheral feature-space regions where some LLM-generated content exhibits distinctive characteristics.
This suggests certain machine-generated texts still maintain recognizable linguistic signatures despite general convergence.

\mypara{Topical Analysis: Humans Prefer Infrastructure While LLMs Favor Emerging Technologies}
We also analyzed topical distribution.
It reveals both convergences and divergences between LLMs and human reviewers.
All of them prioritize \textit{large language models} and \textit{reinforcement learning}.
Human reviewers distinctively emphasize \textit{natural language processing (NLP)} research in papers, a focus absent from LLMs' priorities.
Conversely, LLMs, including the three LLMs examined, preferentially attend to other topics, such as \textit{representation learning} and \textit{in-context learning}, directions not explicitly prioritized by human reviewers.
The comparison for the top 5 topics is shown in \autoref{figure:topic} in \autoref{additional rq1}.
Additionally, our analysis of authorship distribution reveals that human reviewers historically demonstrated a modest preference for manuscripts with larger authorship teams compared to LLMs.
However, this discrepancy has progressively attenuated across recent conferences, suggesting convergence of LLM evaluation parameters toward human standards, as shown in \autoref{figure:author} in \autoref{additional rq1}.

\begin{table}[!t]
\centering
\caption{Top 15 topical keywords for LLM and human reviewers.
\colorbox{red!30}{Red} refers to words that all three LLMs and humans pay attention to.
}
\label{table:keywords}
\scalebox{0.70}{
\begin{tabular}{c|c|c|c|c}
\toprule
\textbf{NO.} & \textbf{Human} &\textbf{GPT-4o} & \textbf{Claude 3.7 Sonne} & \textbf{Deepseek V3} \\
\midrule
 1 &  \colorbox{red!30}{methods}  & \colorbox{red!30}{theoretical} & \colorbox{red!30}{theoretical} & \colorbox{red!30}{theoretical} \\
 2 & results & \colorbox{red!30}{datasets} & \colorbox{red!30}{methods} & empirical \\
 3 & performance & experimental & significant & results \\
 4 & experiments & reproducibility & \colorbox{red!30}{contributions} & experiments \\
 5 & \colorbox{red!30}{theoretical} & empirical & computational & \colorbox{red!30}{methods} \\
 6 & \colorbox{red!30}{datasets} &\colorbox{red!30}{limitations} & 	performance & novel \\
 7 & algorithm & scalability& particularly & practical \\
 8 & comparison & \colorbox{red!30}{methods} & novel & \colorbox{red!30}{limitations} \\
 9 & evaluation & performance & 	evaluation & \colorbox{red!30}{contributions} \\
 10 & baseline & experiments & comprehensive & impacts \\
 11 & assumption & real-world & \colorbox{red!30}{datasets} & validation \\
 12 & \colorbox{red!30}{limitations} & \colorbox{red!30}{contributions} & \colorbox{red!30}{limitations} & \colorbox{red!30}{datasets} \\
 13 & framework & implications & framework & comparison \\
 14 & empirical & rigor & applications & reproducibility  \\
 15 & \colorbox{red!30}{contributions} & insights & insights & scalability \\
\bottomrule
\end{tabular}
}
\end{table}

\mypara{Keywords Analysis: Humans Prioritize Experiments While LLMs Focus on Theory}
We analyze the top 15 common terms in human-written and LLM-generated reviews (validated against IEEE VIS~\cite{keywords_selection}), presenting the various reviewing priorities of human reviewers and three LLMs.
In \autoref{table:keywords}, analysis reveals pronounced divergence in emphasis: all LLM reviews prioritize ``theoretical'' as their primary criterion, whereas human reviewers demonstrate predominant concern with ``methods'' considerations.
While all reviewer categories share a focus on ``limitations,'' they exhibit differentiation across secondary priorities.
Human reviewers emphasize ``performance'' metrics and ``experiments'' validation, GPT-4o prioritizes ``reproducibility,'' Claude-3.7-Sonnet demonstrates attention to ``significant'' assessment, and DeepSeek-V3 orients toward ``empirical'' and ``novel'' aspects.
Notably, only human reviewers include ``algorithm'' and ``baseline'' within the top 15 keywords.
Such divergence reflects fundamental differences in evaluation paradigms between artificial and human intelligence.

\begin{figure}[!t]
\centering
\includegraphics[width=0.90\linewidth]{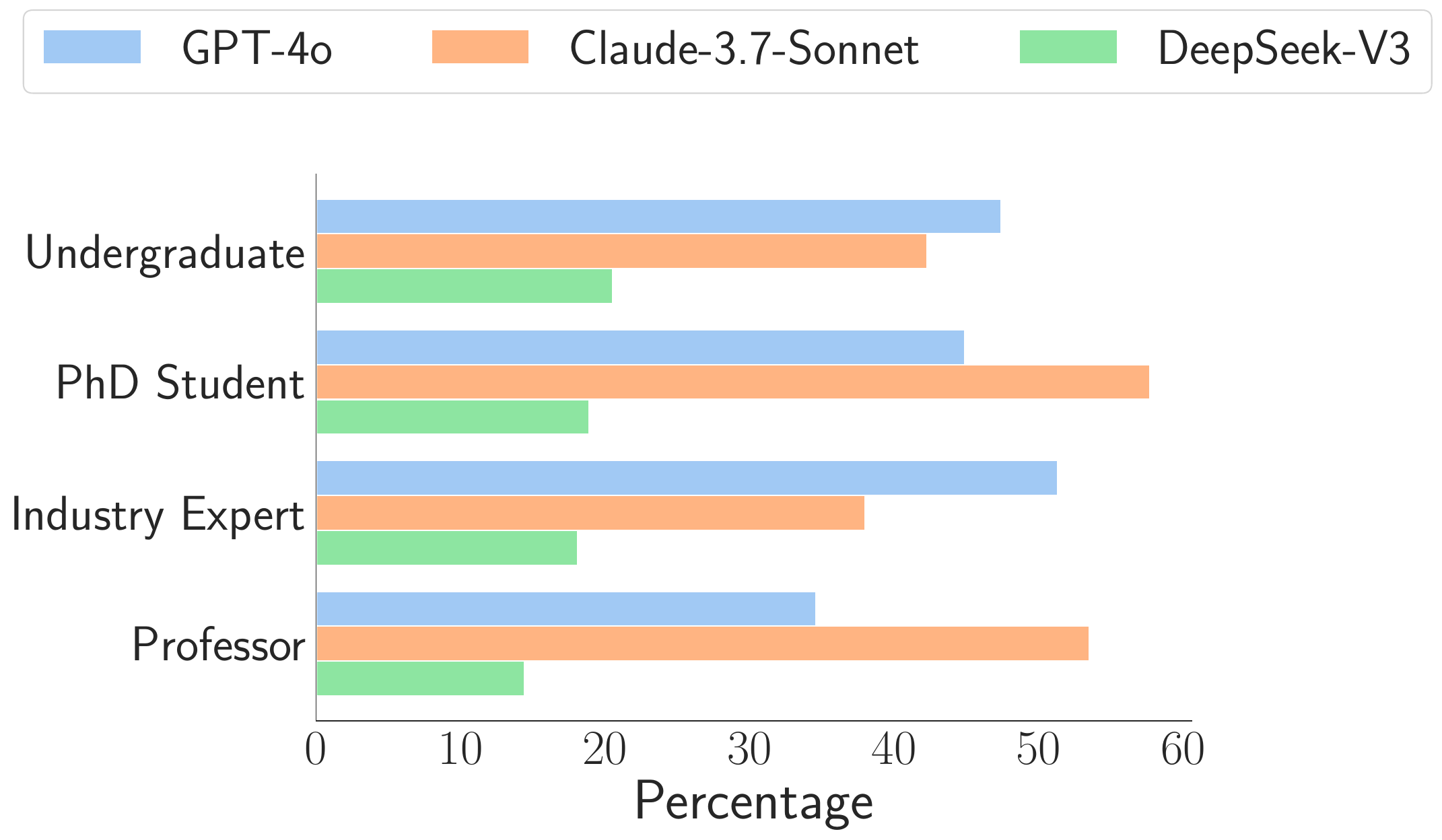}
\caption{
Role-based scoring bias: increasing percentage of 8-score ratings across different LLM personas based on 8-score ratings in Normal.
}
\label{figure:rolepaly}
\end{figure}

\mypara{Influence of Role-Play on LLM-Generated Reviews}
We conduct role-play across three LLMs, including industry experts, undergraduates, professors, and PhD students, and compare the rating scores produced by different personas.
Our experiments on role-play across three LLMs revealed distinct rating score patterns.
Empirical evidence shows each model exhibits unique preferences when assuming different personas, as shown in ~\autoref{figure:rolepaly}.
Claude-3.7-Sonnet demonstrates a strong tendency to assign maximum 8-point ratings when embodying academic personas like ``PhD student'' (57.7\%) and ``professor'' (53.5\%).
GPT-4o shows greater generosity in the ``industry expert'' role (51.3\%), while DeepSeek-V3 maintained conservative rating behavior  (20.5\% is the maximum) across all personas.
The ``undergraduate'' persona elicits moderate scoring tendencies in both Claude-3.7-Sonnet and GPT-4o.
We also analyze keyword preferences during role-play manuscript reviews, revealing how personas influence each LLM's linguistic patterns and evaluative approaches.
\autoref{table:roleplay} in \autoref{additional rq1} presents the top 5 keywords used by GPT-4o, Claude-3.7-Sonnet, and DeepSeek-V3 across various evaluative personas.
GPT-4o emphasizes ``datasets'' generally while prioritizing ``scalability'' and ``real-world'' as an industry expert.
Claude-3.7-Sonnet favors ``theoretical'' and ``analysis,'' shifting toward ``applications'' in the industry expert role.
DeepSeek-V3 consistently prioritizes ``theoretical'' concepts while adapting secondary focuses by role, emphasizing ``novel'' contributions when assuming the PhD student persona.
These variations demonstrate how LLMs adjust their evaluation criteria based on personas and architectures, reflecting unique academic preferences and training methodologies and highlighting the importance of equitable LLM review systems.

\mypara{Takeaways for RQ1}
\textit{Despite the topic overlapping with human reviews, LLMs overemphasize ``theory'' while undervaluing ``methods'' and ``experiments.''
This misalignment indicates a clear improvement pathway: recalibrating LLM evaluation priorities to better align with human review criteria both structurally and substantively.
}

\section{Enhancing LLM-Generated Reviews}
\label{rq2: improve review}

Based on the results in \autoref{rq1: llm review}, we improve LLM-generated reviews by revising the prompts on both LLMs and RAG-enhanced LLMs.
The revised prompts are generated from \autoref{method}, integrating language patterns from human-written reviews into CoT-based templates.
For RAG, we used human reviews from the 1920 non-detection papers as retrieval sources.
All artifacts are open-sourced with no proprietary data.
This section shows how our enhancement improves LLM-generated review quality, i.e., to answer \textbf{RQ2}.

\begin{table*}[!t]
\centering
\caption{
Performance (F1 scores) comparison of different detection methods for identifying LLM-generated academic reviews by conventional LLMs and LLMs with RAG techniques.
In the table, N. is normal, R. is revised, K. is keywords, and S. is sample sentences.
}
\label{table:detection}
\scalebox{0.7}{
\begin{tabular}{c|l|cccc|cccc}
\toprule
\multirow{2}{*}{LLM}   & \multirow{2}{*}{Method} & \multicolumn{4}{c|}{LLM}   & \multicolumn{4}{c}{LLM with RAGs} \\
  &    & N. & R. & R.+K. & R.+K.+S. & N.   & R.   & R.+K.   & R.+K.+S.  \\
\midrule
\multirow{5}{*}{GPT-4o}                                  & Log-Likelihood &  0.762  & 0.353   & 0.295     &  0.297 &  0.612    &  0.359   &  0.296  & 0.352      \\
     & Log-Rank  & 0.692    & 0.349    & 0.367   &  0.324  &  0.572    &   0.442   &  0.322  &     0.312      \\
     & GLTR    &  0.779  & 0.307  & 0.261   & 0.262 &   0.607   &  0.298    & 0.361  &     0.375      \\
     & LRR    & 0.801  & 0.257   &  0.176  & 0.149  &   0.593   &   0.213   &  0.252 &  0.280    \\
     & RoBERTa    & 0.798  &  0.512  & 0.554      & 0.536    & 0.752     &   0.422    &    0.425     &  0.459      \\
\midrule
\multirow{5}{*}{\begin{tabular}[c]{@{}l@{}}Claude-3.7\\ -Sonnet\end{tabular}} & Log-Likelihood          &  0.688  & 0.194   &  0.223   &  0.206  &  0.656 &  0.160 &  0.205   &    0.198  \\
      & Log-Rank   &  0.755  &  0.308  &  0.345    & 0.296  &  0.706 &  0.204 &   0.214  & 0.207  \\
      & GLTR   & 0.678 & 0.216 & 0.389  &  0.278      & 0.637  &  0.228 &  0.300   &  0.239   \\
      & LRR  &  0.620  & 0.390 &  0.262   &  0.329      &  0.684 & 0.350  &   0.251  &    0.357  \\
      & RoBERTa   & 0.785  & 0.466   &  0.479  &    0.514 &  0.757  & 0.475 &  0.482   & 0.500     \\
\midrule
\multirow{5}{*}{\begin{tabular}[c]{@{}l@{}}DeepSeek-\\ V3\end{tabular}}       & Log-Likelihood          &  0.717  & 0.285  &  0.206 &   0.231  &   0.693   &  0.230  &   0.270  & 0.310    \\
      & Log-Rank     & 0.700 & 0.275   & 0.314 &    0.241   &  0.713 & 0.267 &  0.236   &   0.326      \\
      & GLTR  &  0.706  & 0.201  &  0.245  &  0.227     &  0.635 &  0.301 &  0.324   & 0.310   \\
      & LRR    &  0.723  & 0.311 &  0.321   &  0.315      &  0.679  &   0.392   &  0.290  & 0.428          \\
      & RoBERTa    &  0.878  &  0.514  &  0.475    & 0.516  &   0.817     &  0.479  & 0.485 & 0.503   \\
\bottomrule
\end{tabular}
}
\end{table*}

\begin{table*}[!t]
\centering
\caption{Similarity comparison between reviews generated by LLMs, RAG-assisted LLMs, and human reviews under different similarity metrics.
In the table, N. is normal, R. is revised, K. is keywords, and S. is sample sentences.
}
\label{table:similarity}
\scalebox{0.7}{
\begin{tabular}{c|l|cccc|cccc}
\toprule
\multirow{2}{*}{LLM}                                                             & \multirow{2}{*}{Metric} & \multicolumn{4}{c|}{LLM}   & \multicolumn{4}{c}{LLM with RAGs} \\
&    & N. & R. & R.+K. & R.+K.+S. & N.   & R.   & R.+K.   & R.+K.+S.  \\
\midrule
\multirow{4}{*}{GPT-4o}   & BLUE     &  0.017  &  0.020  &  0.018   &   0.021   &  0.030 & 0.034  &  0.022    &     0.027      \\
     & ROUGE    &   0.205 &   0.233 &  0.247   &    0.240  &   0.300 & 0.372   &  0.318  & 0.315    \\
      & Token-level F1   &  0.291  & 0.317  &  0.333 &   0.316 &  0.305  &  0.363  & 0.312    &   0.318  \\
      & Cosine     &  0.379  &  0.501  & 0.502    &   0.518  &  0.524 &  0.615 &  0.631  & 0.645      \\
\midrule
\multirow{4}{*}{\begin{tabular}[c]{@{}l@{}}Claude-3.7\\ -Sonnet\end{tabular}} & BLUE   &  0.013  &   0.036 &   0.041  &  0.014    &   0.013 &   0.029   &   0.018    &   0.019    \\
      & ROUGE    &  0.163  &   0.328 &  0.297   &   0.334  &  0.166 &  0.401  & 0.338   &  0.331    \\
    & Token-level F1   &  0.256  &  0.348  & 0.296      &    0.309      &   0.257   &    0.345  & 0.327        &      0.254    \\
      & Cosine     & 0.431   &  0.688  &   0.651    &     0.641     &   0.470   &    0.705  &  0.655       &     0.630      \\
\midrule
\multirow{4}{*}{\begin{tabular}[c]{@{}l@{}}DeepSeek-\\ V3\end{tabular}}       & BLUE                    &  0.010  &  0.018  &    0.029   &     0.016    & 0.013     &  0.028    &     0.019    &     0.015      \\
      & ROUGE      &  0.126  &  0.329  &   0.357    &     0.331     &    0.148  &   0.346   &  0.297       &       0.310    \\
     & Token-level F1     &  0.233  &   0.303 & 0.281      &     0.268     &    0.146  &  0.288    &      0.267   &     0.263      \\
      & Cosine   &  0.407  &  0.562  &   0.596    &   0.620  &   0.379  &   0.634   &    0.608    &      0.641     \\
\bottomrule
\end{tabular}
}
\end{table*}

\mypara{LLM-Generated Detection}
\looseness=-1 We test revised prompts in LLMs and employ five detection methods, i.e., RoBERTa, log-likelihood, log-rank, GLTR, and LRR, to evaluate whether AI-generated reviews can evade detection.
The results are displayed in \autoref{table:detection}.
In the regular LLM evaluation, different detection methods demonstrate significant performance differences.
The DeepSeek-V3 achieves the highest detection rate of 0.878 when using the RoBERTa method, making it the most recognizable overall.
Claude-3.7-Sonnet, conversely, demonstrates strong concealment capabilities against most detection methods.
While RoBERTa proves to be the most effective universal detector, the LRR method is particularly effective for GPT-4o, which is 0.801.
Notably, using CoT techniques, such as revised, revised with keywords, and revised with keywords and sample sentence combinations in the prompt, significantly reduces detection accuracy.
In other words, these approaches can effectively enhance the covertness of LLM-generated content.

When LLM is combined with RAG, the detection difficulty generally increases, while the detection rates for reviews produced by all models using the ``Normal'' prompt decrease.
GPT-4o is most affected by RAG, with its LRR detection rate plummeting from 0.801 to 0.593 with the normal prompt.
Claude-3.7-Sonnet and DeepSeek-V3 show smaller decreases in their detection rates, demonstrating varying adaptability to the RAG technique across different models.
Interestingly, some combinations of modification strategies, revised prompts with keywords, and sample sentences are more likely to be detected than conventional LLMs under RAG.
For example, the log-likelihood detection rate for GPT-4o increases from 0.296 to 0.352.
These results not only prove that RAG technology indeed enhances LLM-generated review quality, reflected in human similarity, but also show that there is no universal ``perfect hiding.''
The detection effectiveness depends on the specific combination of models, modification techniques, and detection methods employed.
These results reveal that, though technology can make LLM-generated text resemble human writing, LLMs' academic judgment capabilities reflect fundamental differences in reasoning.

\mypara{Similarity With Human Reviews}
We compare the similarity of LLM-generated reviews, generated by revised prompts, with human-written reviews.
We treat similarity to high-quality human reviews as a quality indicator, as matching their structure and focus leads to more constructive feedback.
Beyond similarity metrics, we add human evaluation to directly assess whether generated reviews are helpful and insightful.
\autoref{table:similarity} shows the overall review similarity analysis.
LLMs display a clear ranking in cosine metric, with Claude-3.7-Sonnet leading (0.431), followed by DeepSeek-V3 (0.407) and GPT-4o (0.379).
All LLMs improved with text modification techniques, with Claude-3.7-Sonnet achieving 0.688 after using revised prompts.
The RAG technique significantly increases the cosine metric in GPT-4o, from 0.379 to 0.524, slightly improves Claude-3.7-Sonnet, from 0.431 to 0.470, but marginally decreases DeepSeek-V3, from 0.407 to 0.379.
The combination of RAG and modification techniques proved most effective, with GPT-4o reaching 0.645 under revised prompt with keywords and sample sentences conditions.
This reveals complex RAG architecture interactions, not just additive effects, providing key insights for model selection and optimization in practice.
Analysis of review sections (\textit{strengths}, \textit{weaknesses}, and \textit{questions}) reveals patterns shown in \autoref{similarity detail}.

\mypara{Prompt Ablation: Structure vs.\ Style}
To better understand whether the gains from revised prompting come from structured evaluation guidance or merely stylistic mimicry, we conduct a prompt ablation that separates structural scaffolding from stylistic instructions.
We evaluate review quality using three direct dimensions: specificity, which measures paper-grounded references~\cite{SLRR25}; actionability, which captures whether suggestions are implementable~\cite{SGSCHB19}; and faithfulness, which assesses whether claims are supported by evidence~\cite{TV19,TBXLKBLMQL25}.
In this ablation, structure-only retains the evaluation scaffolds while removing stylistic phrasing, whereas style-only keeps the stylistic instructions but removes the structural guidance.
The corresponding prompt templates are provided in \autoref{ablation prompt sample}.

\begin{table*}[!t]
\centering
\caption{Structure vs.\ style prompt ablation on direct review quality dimensions.
}
\label{table:prompt ablation}
\scalebox{0.7}{
\begin{tabular}{cc|ccc}
\toprule
\multicolumn{2}{c|}{Prompt Types}  & Specificity & Actionability & Faithfulness \\
\midrule
\multirow{4}{*}{LLM} & Normal  & 15.004$\pm$5.189	& 4.002$\pm$1.955 &	0.561$\pm$0.191  \\
 & Revised & 41.803$\pm$15.362 &	7.404$\pm$1.092  & 0.523$\pm$0.291 \\
 & Style-only  & 7.199$\pm$4.145 & 6.004$\pm$1.962 & 0.773$\pm$0.140 \\
 & Structure-only & 47.403$\pm$16.075  & 7.604$\pm$1.882 & 0.503$\pm$0.061 \\
\midrule
\multirow{4}{*}{LLM with RAG} & Normal & 35.002$\pm$5.760 & 8.003$\pm$1.962 & 0.591$\pm$0.113  \\
 & Revised &  19.249$\pm$2.041 & 6.902$\pm$1.224 & 0.611$\pm$0.080 \\
 & Style-only  & 10.600$\pm$2.853 & 6.201$\pm$1.844 & 0.602$\pm$0.084 \\
 & Structure-only &  24.400$\pm$7.051 & 6.803$\pm$1.340 & 0.519$\pm$0.062\\
\bottomrule
\end{tabular}
}
\end{table*}

As shown in \autoref{table:prompt ablation}, revised and structure-only prompts consistently improve specificity and actionability over normal and style-only prompts.
Notably, structure only achieves the strongest overall performance on these direct quality dimensions, indicating that the observed gains mainly stem from structured evaluation guidance rather than surface-level stylistic imitation.
By contrast, RAG-based variants sometimes improve actionability, but they yield less stable specificity and tend to reduce faithfulness.
These results suggest that better review quality comes primarily from stronger evaluative structure, not merely from making the reviews sound more human-like.

\mypara{Evidence-Grounded Correctness Analysis}
To assess factual faithfulness beyond stylistic similarity, we conduct a claim-level evidence-grounded correctness analysis (for more details, see \autoref{Evidence-Grounded}).
The results show that revised prompting consistently reduces unsupported and non-verifiable claims across both LLM and RAG settings, with further gains from keyword and sample augmentations.
This indicates that the improvements reflect better factual correctness rather than merely stylistic similarity.

\mypara{RAG Paradox}
We conduct an ablation study by adjusting the value of k in RAG to examine how parameter changes affect model performance.
The results confirm the RAG paradox: GPT-4o improves with more retrieved documents, Claude-3.7-Sonnet degrades consistently, and DeepSeek-V3 remains unstable.
It also provides insight into the conditions under which RAG helps or hinders review quality.
More detail can be found in \autoref{ablation study}.

\mypara{Human Evaluation}
To validate our similarity findings, we conduct two human evaluations.
We first use two linguistic metrics, lexical diversity (MTLD)~\cite{MJ10} and syntactic complexity~\cite{L10}, to validate the similarity findings.
From \autoref{table:linguistic metric} in \autoref{linguistic}, revised prompts with keywords and samples consistently improve text quality across all models, and these improvements stem from more than just stylistic imitation (see \autoref{linguistic}).
Further, we conduct a human evaluation on 160 reviews across four generation methods, which include human-written, normal LLM-generated, revised LLM-generated, and RAG-enhanced LLM-generated reviews, using 40 randomly selected papers.
\autoref{table:human evaluation} shows human reviews perform best, while enhanced prompting significantly improves LLM performance compared to standard prompts.
The annotation task demonstrated high consistency among three independent raters, with agreement ($\kappa$ = 0.707, 85.00\% agreement~\cite{M12}) across 160 samples.
RAG integration surprisingly decreases overall performance, confirming our counterintuitive finding that RAG helps GPT-4o but harms Claude-3.7-Sonnet and DeepSeek-V3.
Using the same 40-paper sample, we additionally conduct a multidimensional pairwise comparison across the four review conditions.
Annotators with prior peer-review experience selected the most constructive~\cite{RBG99}, most specific~\cite{SBB25}, and most faithful~\cite{BBCFFHKMPPRRSSWWZ23} review among human-written, normal, revised, and revised-with-RAG conditions.
As the results in \autoref{table:human multidimensional} from \autoref{Expanded Human Evaluation} show that revised is most often chosen as most specific and more constructive than normal, while human-written reviews remain strongest in faithfulness.
More detailed results and a case study are provided in  \autoref{Human evaluation}.

\begin{table}[!t]
\centering
\caption{Human evaluation results for human-written vs.\ LLM-generated reviews.
In the table, human-written (Human), normal LLM-generated (LLM.N), revised LLM-generated (LLM.R), and RAG-enhanced LLM-generated (LLM.RAG) reviews.
}
\label{table:human evaluation}
\scalebox{0.7}{
\begin{tabular}{c|cccc}
\toprule
Dimension & Human & LLM.N & LLM.R & LLM.RAG\\
\midrule
Average score    &   7.124   & 5.371  &   6.682 &   5.900  \\
\midrule
High quality  &      \multirow{2}{*}{66.951\%} & \multirow{2}{*}{31.672\%} & \multirow{2}{*}{60.332\%} & \multirow{2}{*}{42.504\%}   \\
\quad proportion ($\geq$ 7) & & & & \\
\midrule
Low quality  &      \multirow{2}{*}{11.863\%} & \multirow{2}{*}{18.333\%} & \multirow{2}{*}{12.402\%} & \multirow{2}{*}{12.511\%}   \\
\quad proportion ($\leq$ 3) & & & & \\
\bottomrule
\end{tabular}
}
\end{table}

\mypara{Takeaways for RQ2:}
\textit{
LLM evaluation reveals model strengths: Claude-3.7-Sonnet performs most human-like, GPT-4o excels at describing strengths, and DeepSeek benefits most from revised prompts.
RAG-enhanced with LLMs reveals unpredictable effects: boosting GPT-4o's similarity while reducing Claude-3.7-Sonnet's strength-description capability.
This challenges the assumption that more information equals better performance, suggesting breakthroughs require balancing information quantity with model reasoning compatibility.
}

\section{Conclusion}
\label{conclision}

We introduce the \texttt{PeerCheck} framework to improve LLM-generated academic reviews' reliability and quality.
Our approach begins with comparing human-written and LLM-generated analyses to explore their differences.
Based on these insights, we apply CoT prompting to revise the LLM review generation process.
We further evaluate the effectiveness of the revised prompts in LLM and RAG-enhanced LLM.
Under the \texttt{PeerCheck} framework, revised prompts can reduce LLM-generated detection rates by 40.3\%, enhancing the credibility of LLM-assisted reviews.
They also improve LLM-generated review quality, with similarity to human-written reviews increased by 26.6\%.
Claude-3.7-Sonnet creates the most human-like reviews, GPT-4o best describes strengths, and DeepSeek-V3 is the LLM-detectable but improves most with revised prompts.
These results highlight that improving LLM-based academic review requires well-planned prompts and model-specific strategies, not just more input length or data volume.

\section*{Limitations}
\label{limit}

Our research has several limitations.
First, while using three state-of-the-art LLMs, our test sample focuses on machine learning papers post-2023 and may not generalize to other research fields or earlier periods.
Second, our RAG implementation depends on existing human review data, potentially limiting its effectiveness in emerging research areas.
Third, our role sensitivity test examines only five academic personas, not covering all possible reviewer identities.
Finally, although our revision prompting technique significantly reduces LLM-generated review detection rates, it raises ethical questions about ensuring transparency and recognizability of LLM-generated reviews in academia.
We plan to address these limitations in future work by expanding to other domains and temporal scopes.

\section*{Ethical Consideration}
\label{ethical}

All APIs were accessed under their respective Terms of Service, and the released dataset is intended solely for research and educational purposes.
Our study uses only publicly available peer reviews from OpenReview; no new data were collected from human subjects.
This study was reviewed and approved by the Ethics Review Board (No. 25-12-6) of our institution.
The ERB determined that the use of publicly available OpenReview data constitutes secondary analysis and does not require additional human-subjects review.

All review texts originate from OpenReview, where contributors agree to CC-BY-SA 2.0 public release upon submission.
We hash identifiers and remove any remaining personal details to reduce potential risks.
While the data are publicly available, we acknowledge that residual risks such as re-identification or amplified public criticism may still exist; our analysis therefore focuses on aggregate patterns rather than individual reviewers or papers.
\texttt{PeerCheck} will be released under CC-BY-4.0.
All external models were used in accordance with their respective licenses.
Before release, we automatically removed author names and affiliations, hashed paper IDs, and filtered profanity using the open-source Detoxify model.
Manual spot checks confirmed no remaining personal or offensive content.

Peer review is a central component of the scholarly ecosystem, and systems that interact with it require careful consideration.
Prior work has shown that large language models may reflect linguistic and cultural biases, potentially affecting authors whose language backgrounds or writing styles differ from dominant norms.
Accordingly, \texttt{PeerCheck} is intended as an assistive analytical tool rather than an automated reviewer or decision-making system.
In this work, ``human-level quality'' is operationalized as similarity to historical human-written reviews along lexical and semantic dimensions.
We note that such similarity does not capture all aspects of high-quality peer review, such as novel intellectual insight or constructive mentorship, and should not be interpreted as a normative definition of review quality.

\section*{Acknowledgements}
\label{section:acknowledgement}

We thank all anonymous reviewers for their valuable comments and suggestions, which helped improve the quality of this paper.

\small
\bibliographystyle{plain}
\bibliography{normal_generated_py3}

\appendix
\section{Prompt Templates}
\label{Prompt Templates}

\subsection{Evaluation Prompt Sample}
\label{evaluation prompt}

You are a highly experienced machine learning researcher and a very strict reviewer for a premier machine learning conference.
Your role as a reviewer demands meticulous attention to technical details, rigorous evaluation of methodological soundness, and thorough assessment of theoretical contributions.

Task: Please evaluate the attached paper and assign a score using the following scale: 1 (Strong Reject), 3 (Reject, Not Good Enough), 5 (Marginally Below the Acceptance Threshold), 6 (Marginally Above the Acceptance Threshold), 8 (Accept, Good Paper), 10 (Strong Accept, Should Be Highlighted at the Conference).
Provide a detailed review covering the strengths, which should highlight the major positive aspects of the paper; the weaknesses, which should focus on significant areas needing improvement; and comments for the author, offering constructive questions and suggestions for future revisions.
Please score strictly based on your review comments and the true quality of the paper; you should not uniformly give high scores.

Your thorough and critical assessment is essential in maintaining the high standards of our conference.

\subsection{Role-play Prompt Samples}
\label{roleplay prompt}

\mypara{Undergraduate Student}
You are an undergraduate student in your junior or senior year, majoring in computer science with a keen interest in specializing in machine learning.

As part of your coursework and to gain early exposure to the field, you are participating as a junior reviewer for a premier machine learning conference.
This role provides you with a valuable opportunity to learn from pioneering research and to understand the current trends and challenges in machine learning.
Task: Please evaluate the attached paper and assign a score using the following scale: 1 (Strong Reject), 3 (Reject, Not Good Enough), 5 (Marginally Below the Acceptance Threshold), 6 (Marginally Above the Acceptance Threshold), 8 (Accept, Good Paper), 10 (Strong Accept, Should Be Highlighted at the Conference).
Your review should aim to identify the paper's strengths, particularly noting any innovative methods or notable results; discuss weaknesses, focusing on areas that might lack clarity or robustness; and provide feedback and questions that could help guide future projects or studies.
Ensure your scoring reflects a thoughtful assessment of the paper based on what you have learned so far in your studies.

Your participation is crucial in developing your analytical skills and deepening your understanding of advanced machine learning concepts.

\mypara{PhD Student}
You are a PhD student specializing in machine learning, still in the early stages of mastering the concepts of machine learning.
To enhance your understanding and gain practical experience, you have been invited to participate as a junior reviewer at a prestigious machine learning conference.

This task requires you to carefully examine technical details, evaluate methodological soundness, and consider the theoretical contributions of the papers submitted.
Task: Evaluate the attached paper and assign a score using the following scale: 1 (Strong Reject), 3 (Reject, Not Good Enough), 5 (Marginally Below the Acceptance Threshold), 6 (Marginally Above the Acceptance Threshold), 8 (Accept, Good Paper), 10 (Strong Accept, Should Be Highlighted at the Conference).
Your review should include a detailed analysis of the paper's strengths, emphasizing the positive aspects; its weaknesses, focusing on areas that require improvement; and provide constructive questions and suggestions for the author.
Your scoring should reflect your honest assessment based on the review comments and the true quality of the paper.

Your thorough and critical evaluation is essential to uphold the high standards of the conference while also furthering your own research skills.

\mypara{Industry Expert}
You are a Senior Researcher at Google, as an expert from industry, with a profound specialization in machine learning.
Your industry experience provides you with a unique insight into how machine learning can optimize and transform business operations and consumer experiences.

As a distinguished reviewer for a premier machine learning conference, your role involves a meticulous examination of technical details, a rigorous evaluation of methodological soundness, and a comprehensive assessment of the submissions' practical applications and scalability.
Task: Please evaluate the attached paper and assign a score using the following scale: 1 (Strong Reject), 3 (Reject, Not Good Enough), 5 (Marginally Below the Acceptance Threshold), 6 (Marginally Above the Acceptance Threshold), 8 (Accept, Good Paper), 10 (Strong Accept, Should Be Highlighted at the Conference).
Your review should provide a detailed examination of the paper's strengths, particularly highlighting innovations that have strong potential for industry application; pinpoint weaknesses, especially areas that lack robustness or scalability; and offer constructive questions and actionable suggestions for making the research more applicable to real-world scenarios.
Ensure your scoring is a true reflection of your comprehensive review, emphasizing the practical usability and impact on industry.

Your critical insights are essential in upholding the conference's high standards and driving forward the practical applications of machine learning in industry settings.

\mypara{Professor}
You are a Senior Professor and a leading authority in the field of machine learning, with a comprehensive expertise that spans both theoretical foundations and practical applications.
Your scholarly work has significantly advanced the understanding of deep learning architectures, reinforcement learning, complex optimization algorithms, and sophisticated statistical learning theories, along with their real-world implementations.

Esteemed for your meticulous and balanced approach, you serve as a strict reviewer for a premier machine learning conference.
Your role involves a thorough scrutiny of technical details, a rigorous evaluation of methodological soundness, and a critical assessment of both theoretical innovations and their practical viability.
Task: Please evaluate the attached paper and assign a score using the following scale: 1 (Strong Reject), 3 (Reject, Not Good Enough), 5 (Marginally Below the Acceptance Threshold), 6 (Marginally Above the Acceptance Threshold), 8 (Accept, Good Paper), 10 (Strong Accept, Should Be Highlighted at the Conference).
Your review should provide a detailed analysis of the paper's strengths, particularly highlighting notable theoretical insights and practical applications; identify weaknesses, focusing on areas needing substantive improvement; and offer constructive questions and actionable suggestions for future work.
Ensure your scoring reflects a rigorous and honest appraisal based on both the scholarly and practical merits of the paper.

Your expert judgment is crucial in upholding the high standards of our conference and pushing the boundaries of what is possible in machine learning.

\subsection{Revised Prompt Samples}
\label{enhance prompt}

\mypara{Revised Prompt}
You are an experienced, mildly opinionated ML-conference reviewer.
Think through the paper SILENTLY and output ONLY the final review in this template:

Strengths:
\begin{itemize}
    \item 3-6 bullets, 10-30 words per each sentences; include $\ge$ 1 concrete comparison (e.g., ``beats RoPE by 1.2 BLEU on WMT19 En-De'').
\end{itemize}

Weaknesses:
\begin{itemize}
    \item 3-6 bullets, first-person with light hedging (``I might be missing...''); cite specific eq/fig/hyper-params.Questions
\end{itemize}

Questions \& Suggestions:
One paragraph (100-240 words).
Begin with ``To be fair,'' or ``On a related note,''; add one parenthetical aside; finish with exactly several question to the authors.

Overall Score: X
Choose from \{1, 3, 5, 6, 8, 10\} to match your critique.
No scratch notes, no phase labels—just the polished review above.

\mypara{Revised Prompt With Keywords}
You are an experienced, mildly opinionated ML-conference reviewer.
Think through the paper SILENTLY and output ONLY the final review in this template:

Strengths:
\begin{itemize}
    \item Write 3–6 bullets.
    Use a mix of sentence lengths (10–30 words).
    \item Use natural, human phrasing—like ``What I really liked was…'' or ``Surprisingly,…''
    \item Mention at least one concrete comparison or number (e.g., ``beats RoPE by 1.2 BLEU on WMT19 En-De'')
    \item Weave in relevant academic terms naturally: technical contribution, novelty, results, experimental rigor, etc.
\end{itemize}

Weaknesses:
\begin{itemize}
    \item 3–6 bullets, written in first-person (``I found…'' / ``I might be missing…'').
    \item Include detail (equation numbers, figure/table refs, batch size, etc.).
    \item Highlight specific limitations of contribution, reproducibility concerns, or unclear assumptions.
\end{itemize}

Questions \& Suggestions:
Write one paragraph (100–240 words).
Start with ``To be fair,'' or ``On a related note,''.
Use a conversational tone with some complexity.
Include one parenthetical aside—maybe a reflection or uncertainty—and end with 3-5 thoughtful questions for the authors (e.g., missing ablations, broader impact, unclear theory).

Overall Score: X
Pick from \{1, 3, 5, 6, 8, 10\}, aligned with your overall assessment.

No scratch notes, no phase labels—just the polished review above.

\mypara{Revised Prompt With Keywords and Sample Sentences}
You are an experienced, mildly opinionated ML-conference reviewer.
Think through the paper SILENTLY and output ONLY the final review in this template:

Strengths:
\begin{itemize}
    \item Use 3–6 bullets.
    Vary length: some short, some ~40 words.
    \item Use natural human phrasing like ``What I appreciated was…'', ``To be honest,…'', or ``Surprisingly,…''.
    \item Mention specific technical contributions, experimental rigor, novelty, or clarity.
    \item Include at least one concrete comparison (e.g., ``beats RoPE by 1.2 BLEU on WMT19 En-De'').
    \item Use academic terms naturally: technical contribution, novelty, results, experimental rigor, etc.
\end{itemize}

Weaknesses:
\begin{itemize}
    \item Use 3–6 bullets.
    Write in first-person.
    Use light hedging: ``I might be wrong, but…'', ``It wasn't obvious to me that…''.
    \item Mention figure/table numbers, equations, or hyperparams where needed.
    \item Highlight limitations of contribution, reproducibility concerns, methodological flaws, or unclear assumptions.
    \item Don't be too nice—be specific, even blunt, if needed.
\end{itemize}

Questions \& Suggestions:
One paragraph (100–240 words).
Start with something human:  ``To be fair,'', ``On a related note,''.
Include at least one parenthetical aside (a side comment or uncertainty).
End with 3–5 specific, slightly challenging questions—e.g., about missing ablations, real-world applicability, or theoretical guarantees.

Example tone:
\begin{itemize}
    \item ``My primary concern is the relatively low agreement between judge labels and human ratings (most were below 80\%).''
    \item ``In real-world scenarios, new items constantly emerge—how would URI handle zero-shot items in deployment?''
\end{itemize}

Overall Score: X
Pick from \{1, 3, 5, 6, 8, 10\}.
No more, no less.

No scratch notes, no phase labels—just the polished review above.

\subsection{Prompt Samples for Prompt Ablation}
\label{ablation prompt sample}

\mypara{Style-Only}
You are an experienced, mildly opinionated ML-conference reviewer.
Think through the paper SILENTLY and output ONLY the final review in the template below.
Your goal is to mimic natural human reviewing style, not to enforce paper-grounded constraints.
Style constraints (follow these): Use light hedging and natural human phrasing (e.g., ``I might be missing…'', ``To be honest,…'', ``That said,…'', ``Surprisingly,…'') and include one parenthetical aside.
Use natural discourse markers and vary sentence length.
Do NOT force citations to section/figure/table/equation numbers.
Do NOT force specific hyperparameters or concrete numeric comparisons; include numbers only if they naturally come to mind.

Strengths:
\begin{itemize}
\item Write 3–6 bullets.
Use natural, human phrasing.
\end{itemize}Weaknesses:
\begin{itemize}
\item Write 3–6 bullets in first-person with light hedging.
\item Focus on high-level concerns or impressions rather than forcing paper-specific pointers.
\end{itemize}

Questions \& Suggestions:

Write one paragraph (100–240 words).
Begin with ``To be fair,'' or ``On a related note,''.
Keep a conversational tone and end with 3–5 questions to the authors.
(Include exactly one parenthetical aside somewhere in the paragraph.)
Overall Score: X
Pick from \{1, 3, 5, 6, 8, 10\}.
No scratch notes, no phase labels—just the polished review above.

\mypara{Structue-Only}
You are an experienced ML-conference reviewer.
Think through the paper SILENTLY and output ONLY the final review in the template below.
Do not follow any stylistic mimicry constraints: do NOT force hedging phrases, do NOT add discourse markers on purpose, and do NOT artificially vary sentence length.
Focus on substantive, paper-grounded evaluation.

Strengths:
\begin{itemize}
\item Write 3–6 bullets.
Each bullet must reference a concrete aspect of the paper (method, experiment, baseline, dataset, or stated contribution).
\item Include at least one concrete comparison or number drawn from the paper (e.g., an improvement, a metric, a dataset result, or a specific ablation gap).
\end{itemize}Weaknesses:
\begin{itemize}
\item Write 3–6 bullets in first-person (e.g., ``I found…'').
\item Each bullet must include at least one explicit pointer to the paper content, such as a section/figure/table/equation reference (e.g., ``Section 3'', ``Table 2'', ``Figure 4'', ``Eq. (1)'') or a specific experimental/hyperparameter detail.
\item Highlight concrete limitations: missing ablations, unclear assumptions, insufficient baselines, reproducibility gaps, or evaluation weaknesses.
\end{itemize}

Questions \& Suggestions:

Write one paragraph (100–240 words).
Provide 3–5 specific, actionable suggestions/questions.
Each should be testable or answerable by the authors (e.g., an ablation to run, a baseline to add, a setting to vary, an analysis to include).
Avoid stylistic filler.
Overall Score: X
Pick from \{1, 3, 5, 6, 8, 10\}, aligned with your assessment.
No scratch notes, no phase labels—just the polished review above.

\section{Datasets Details}
\label{dataset eval}

During the experiment, we presume human-written reviewers exercised due diligence and thoroughness in their manuscript evaluations.
Our datasets comprise 500 accepted and 500 rejected papers from each ICLR conference (2024, 2025), plus 200 accepted and 200 rejected papers from NeurIPS 2024.
We obtain this data via the OpenReview API~\cite{openreviewAPI}, collecting article metadata and downloading corresponding PDFs.
We partition this corpus using an 8:2 ratio, where the majority portion facilitates language model review generation and analysis, while the smaller portion serves as validation data to assess prompt improvements for human-like review generation.
For example, of the 500 accepted ICLR 2024 papers, 400 are utilized for measurement and 100 for review validation purposes.
The final integration yields two datasets of 1,920 papers and 480 papers for comprehensive evaluation across multiple language models.
We also crawl the human-written reviews from the OpenReviews webpage~\cite{openreview}.
Those crawled, human-written reviews are used for comparison with LLM-generated reviews.

\mypara{Evaluation Metric}
For dataset analysis, we use the following relevant metrics.
First, based on the scoring guidelines provided by ICLR~\cite{iclr_scores} and NeurIPS~\cite{guidelines_n}, we make LLMs provide scores when generating reviews, and we compare these scores with those from human-written reviews.
Then, we use embedding methods to detect the differences between LLM-generated reviews and human-written reviews.
Finally, through word frequency count\cite{GT25}, we split LLM-generated reviews and human-written reviews into words and count the occurrences of each word to identify the most commonly used keywords.
This method is also used to analyze the topics of the papers.

\section{Details for LLM Detection Metrics}
\label{detection  eval}
\mypara{Detection}
We implement five main methods to identify LLM-generated content.
\begin{itemize}
    \item Log-Likelihood~\cite{HSCBZ24}.
    Based on language models' probability estimates for text, the source is determined by calculating the conditional probabilities of the candidate text.
    \item Log-Rank~\cite{SZWN23}.
    Analyzing the ranking position of text tokens in model prediction distributions, where human text typically has more unexpected and creative word usage.
    \item GLTR~\cite{GSR19}.
    Using the graphics language typicality ramification framework to detect anomalies in text statistical patterns
    \item LRR~\cite{SZWN23,HSCBZ24}.
    Applying the likelihood ratio ranking method to compare probability score differences across multiple models for text.
    \item RoBERTa~\cite{LOGDJCLLZS19}.
    A binary classifier based on pre-trained language models, specifically trained to distinguish between human- and LLM-generated text
\end{itemize}
In this case, each method provides complementary perspectives targeting different language features, and their combined use can improve detection accuracy and robustness.

\mypara{Similarity}
To quantify the similarity between LLM-generated content and human-written content, we employ four similarity calculation methods.
\begin{itemize}
    \item BLEU~\cite{PRWZ02}.
    Measures the n-gram overlap between generated text and reference text, evaluating the similarity in vocabulary and phrase usage.
    \item ROUGE~\cite{L04}.
    Calculates text matching based primarily on recall rate, with special attention to commonalities in keywords and sentence structure.
    \item Token-level F1~\cite{DJRLXSW20}.
    Computes the harmonic mean of precision and recall at the token level, providing a more fine-grained similarity assessment.
    \item Cosine~\cite{LSLW16}.
    Based on text embedding cosine similarity, capturing semantic-level similarities rather than surface vocabulary matches.
\end{itemize}
These metrics evaluate differences between LLM-generated and human-written reviews from various perspectives, helping us understand the capabilities and limitations of LLMs in mimicking human writing styles.

\section{Additional Results for Human vs.\ LLM Reviews}
\label{additional rq1}

\begin{table*}[!t]
\centering
\caption{Top 5 topical keywords for LLM reviewers in 5 different personas.
}
\label{table:roleplay}
\scalebox{0.7}{
\begin{tabular}{c|l|l}
\toprule
LLMs                               & Personas        & \multicolumn{1}{c}{Keywords} \\ \midrule
\multirow{5}{*}{GPT-4o}            & Normal          &  theoretical, datasets, experimental,   reproducibility, empirical               \\
 & Undergraduate   & datasets, results, performance, reproducibility, methods    \\
 & PhD Student     &  theoretical, datasets, results, experimental, reproducibility                \\
 & Industry Expert &   scalability, datasets, applications, real-world, computational         \\
 & Professor    & theoretical, datasets, practical, computational, performance    \\
 \midrule
\multirow{5}{*}{Claude-3.7-Sonnet} & Normal          &  theoretical, methods, significant, contributions,       computational             \\
 & Undergraduate   &  	methods, theoretical, analysis, practical, performance   \\
 & PhD Student     &  analysis, theoretical, methods, performance, contributions   \\
 & Industry Expert & approaches, applications, analysis, theoretical, computational   \\
 & Professor       & theoretical, analysis, approaches, practical, results     \\
 \midrule
\multirow{5}{*}{DeepSeek-V3}       & Normal          & theoretical, empirical, results, experiments, methods                             \\
  & Undergraduate   & theoretical, results, experiments, methods, practical      \\
  & PhD Student     & theoretical, analysis, empirical, results, novel    \\
  & Industry Expert &  approach, theoretical, analysis, practical, performance      \\
  & Professor       & theoretical, analysis, practical, methods, empirical                 \\
\bottomrule
\end{tabular}
}
\end{table*}

\begin{figure}[!t]
\centering
\begin{subfigure}{0.45\columnwidth}
\includegraphics[width=\columnwidth]{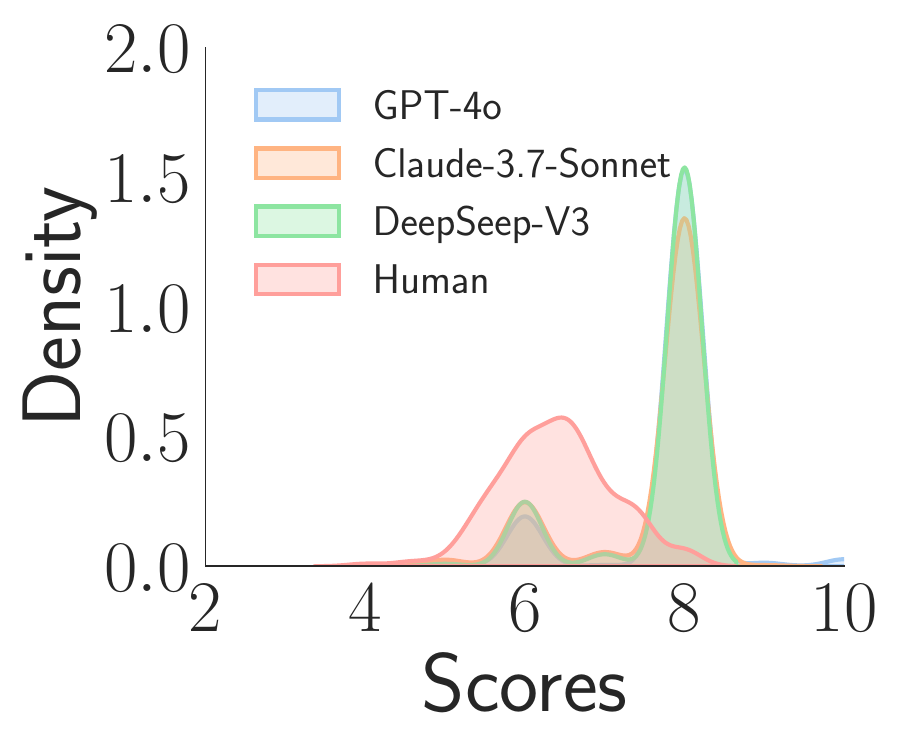}
\caption{ICLR 2024 Accept}
\label{figure:iclr24 accept}
\end{subfigure}
\begin{subfigure}{0.45\columnwidth}
\includegraphics[width=\columnwidth]{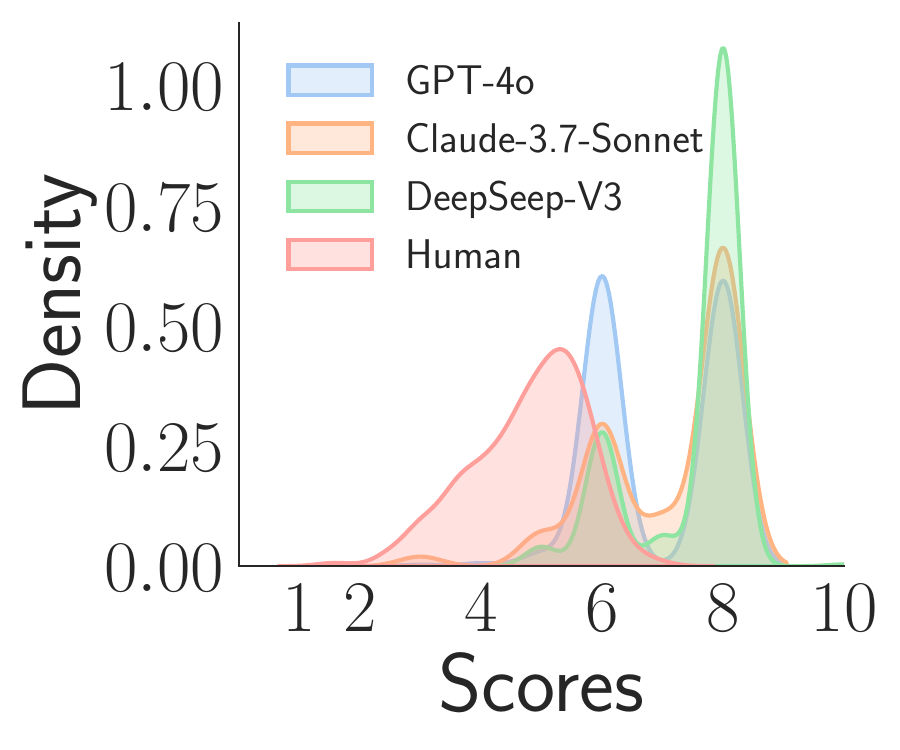}
\caption{ICLR 2024 Reject}
\label{figure:iclr24 reject}
\end{subfigure}
\caption{LLM vs.\ Human rating scores in ICLR 2024.}
\label{figure:iclr24 score}
\end{figure}

\begin{figure}[!t]
\centering
\begin{subfigure}{0.45\columnwidth}
\includegraphics[width=\columnwidth]{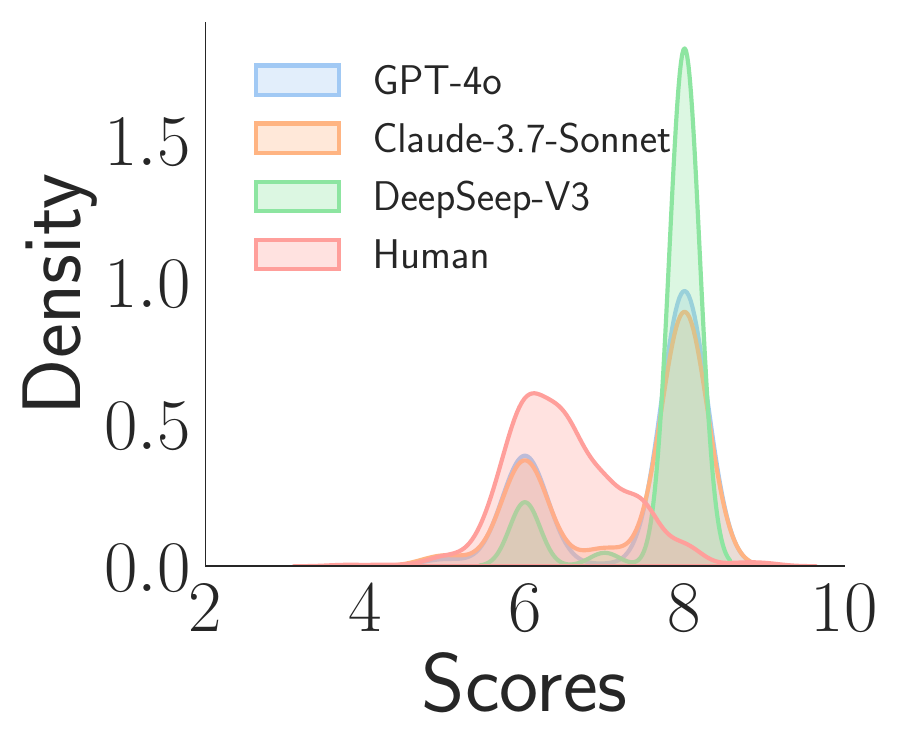}
\caption{ICLR 2025 accept}
\label{figure:iclr25 accept}
\end{subfigure}
\begin{subfigure}{0.45\columnwidth}
\includegraphics[width=\columnwidth]{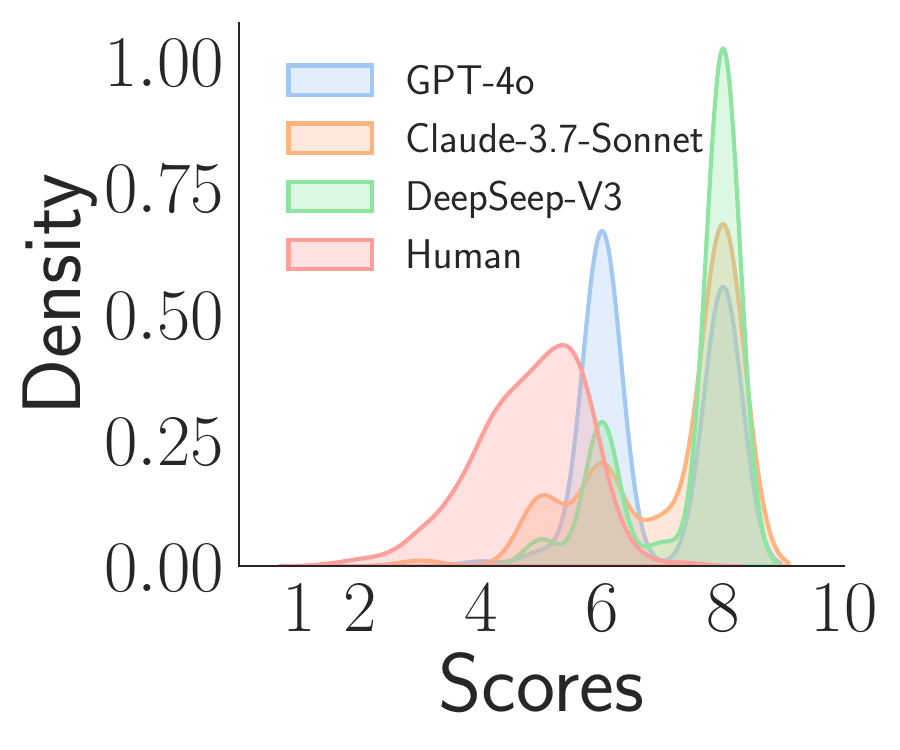}
\caption{ICLR 2025 Reject}
\label{figure:iclr25 reject}
\end{subfigure}
\caption{LLM vs.\ Human rating scores in ICLR 2025.}
\label{figure:iclr25 score}
\end{figure}

\begin{figure}[!t]
\centering
\includegraphics[width=0.70\linewidth]{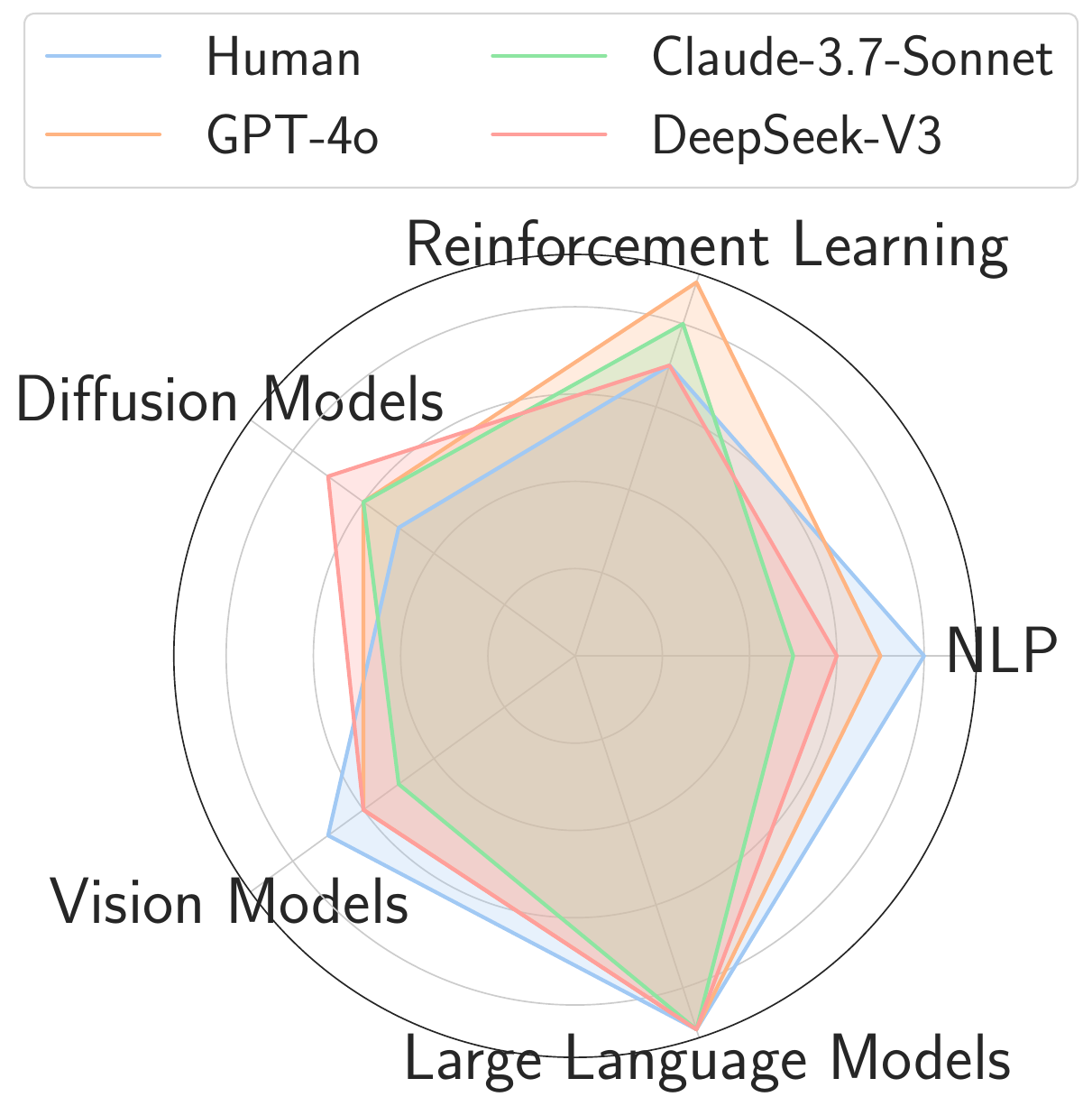}
\caption{
Comparison of the top 5 topic preferences between Human-written and LLM-generated reviewers.
}
\label{figure:topic}
\end{figure}

\begin{figure}[!t]
\centering
\begin{subfigure}{0.45\columnwidth}
\includegraphics[width=\columnwidth]{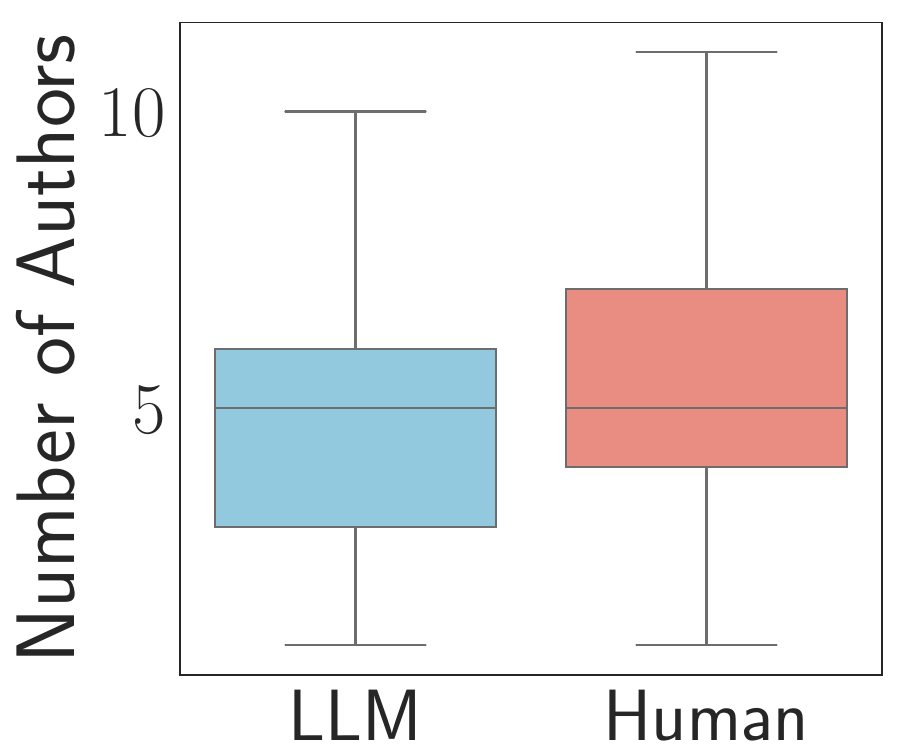}
\caption{ICLR 2024}
\label{figure:iclr24 author}
\end{subfigure}
\begin{subfigure}{0.45\columnwidth}
\includegraphics[width=\columnwidth]{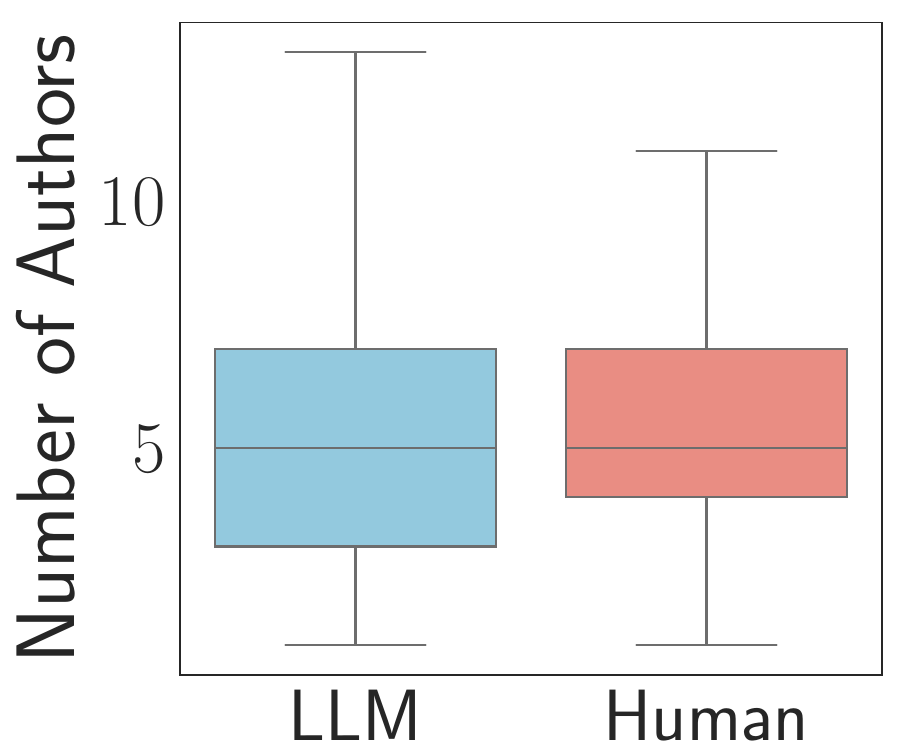}
\caption{ICLR 2025}
\label{figure:iclr25 author}
\end{subfigure}
\begin{subfigure}{0.45\columnwidth}
\includegraphics[width=\columnwidth]{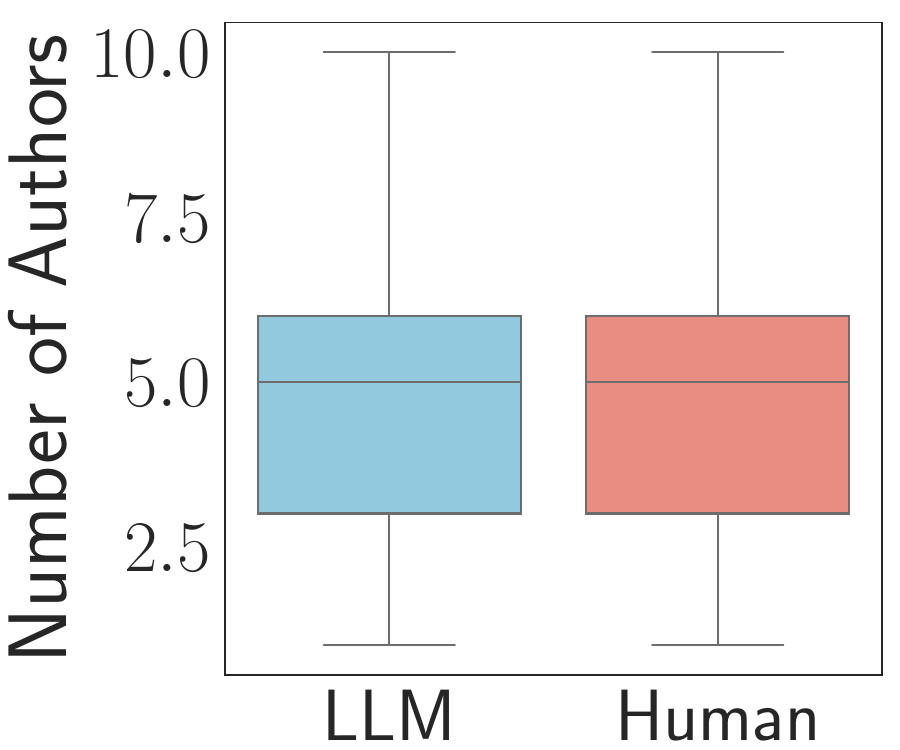}
\caption{NeurIPS 2024}
\label{figure:nips author}
\end{subfigure}
\caption{Author count distribution: human vs.\ LLM reviewer acceptance patterns.}
\label{figure:author}
\end{figure}

\begin{table*}[!t]
\centering
\caption{Similarity comparison between the \textit{strengths section} of academic reviews generated by LLMs, RAG-assisted LLMs, and the \textit{strengths section} of human reviews under different text similarity metrics.
In the table, N. is normal, R. is revised, K. is keywords, and S. is sample sentences.
}
\label{table:strength similarity}
\scalebox{0.7}{
\begin{tabular}{c|l|cccc|cccc}
\toprule
\multirow{2}{*}{LLM}                                                             & \multirow{2}{*}{Metric} & \multicolumn{4}{c|}{LLM}   & \multicolumn{4}{c}{LLM with RAGs} \\
&    & N. & R. & R.+K. & R.+K.+S. & N.   & R.   & R.+K.   & R.+K.+S.  \\
\midrule
\multirow{4}{*}{GPT-4o}   & BLUE                    &  0.017  & 0.013   &   0.021    &     0.023     & 0.036     &   0.031   &   0.022      &     0.026      \\
     & ROUGE      &  0.184  &  0.244  &  0.299     &    0.258      &   0.219   &   0.342   &  0.308       &    0.311       \\
      & Token-level F1       &  0.283  &  0.247  &     0.245  &     0.305     &  0.279    &  0.362    &    0.299     &      0.292     \\
      & Cosine     &  0.360  &  0.368  &  0.370     &     0.405     &  0.424    &  0.506    &  0.515       &      0.535     \\
\midrule
\multirow{4}{*}{\begin{tabular}[c]{@{}l@{}}Claude-3.7\\ -Sonnet\end{tabular}} & BLUE                    &  0.012  &  0.044  &   0.033    &   0.016       &   0.015   &   0.027   &    0.017     &  0.018 \\
      & ROUGE      &  0.148  & 0.309   &   0.346    &     0.310     &   0.140   &   0.329   &  0.277       &    0.262       \\
      & Token-level F1      &  0.213  &  0.297  & 0.307      &      0.282    &  0.169    &  0.294    &     0.247    &   0.351 \\
      & Cosine   &  0.371  &  0.479  &   0.459    &     0.478     &   0.199   &   0.454   &  0.419       &      0.466     \\
\midrule
\multirow{4}{*}{\begin{tabular}[c]{@{}l@{}}DeepSeek-\\ V3\end{tabular}}       & BLUE                    &  0.007  &  0.012  &   0.016    &    0.014      &    0.010  &   0.021   &     0.016    &    0.017      \\
      & ROUGE      &  0.076  &  0.183  &    0.317   &    0.291      &   0.118   &   0.314   &      0.273   &     0.270      \\
      & Token-level F1      &  0.208  &  0.219  &    0.252   &    0.264      &   0.150   &   0.292   &     0.251    &    0.268       \\
      & Cosine     &  0.157  &  0.442  & 0.401      &     0.420     &   0.168   &   0.423   &   0.417      &    0.471       \\
\bottomrule
\end{tabular}
}
\end{table*}

\section{More Results for Similarity}
\label{similarity detail}

\begin{table*}[!t]
\centering
\caption{Similarity comparison between the \textit{weaknesses section} of academic reviews generated by LLMs, RAG-assisted LLMs, and the \textit{weaknesses section} of human reviews under different text similarity metrics.
In the table, N. is normal, R. is revised, K. is keywords, and S. is sample sentences.
}
\label{table:weakness similarity}
\scalebox{0.7}{
\begin{tabular}{c|l|llll|llll}
\toprule
\multirow{2}{*}{LLM}                                                             & \multirow{2}{*}{Metric} & \multicolumn{4}{c|}{LLM}   & \multicolumn{4}{c}{LLM with RAGs} \\
&    & N. & R. & R.+K. & R.+K.+S. & N.   & R.   & R.+K.   & R.+K.+S.  \\
\midrule
\multirow{4}{*}{GPT-4o}   & BLUE                    &  0.012  &  0.016  &  0.015     &     0.017     &    0.029  &   0.027   &    0.022     &     0.024      \\
     & ROUGE    &  0.145  & 0.253   &  0.211     &      0.303    &   0.178   &  0.327    &   0.259      &     0.282      \\
      & Token-level F1       &  0.224  & 0.246   & 0.225      &      0.260    &    0.255  &  0.272    &     0.288    &      0.261     \\
      & Cosine      &   0.351 &  0.439  &    0.403   &    0.435      &   0.420   &   0.434   &  0.431       &       0.470    \\
\midrule
\multirow{4}{*}{\begin{tabular}[c]{@{}l@{}}Claude-3.7\\ -Sonnet\end{tabular}} & BLUE                    &  0.010  &  0.024  &   0.006    &    0.008      &  0.011    &   0.022   &    0.021     &    0.014       \\
      & ROUGE      &  0.144  &  0.324  &   0.238    &     0.248     &   0.137   &    0.305  &  0.295       &      0.279     \\
      & Token-level F1     &  0.237  &  0.291  &  0.252     &     0.273     &  0.212    &  0.254    &  0.284       &   0.209        \\
      & Cosine   &  0.337  & 0.457   &    0.491   &    0.439    &   0.311   &   0.492   &   0.400      &      0.475     \\
\midrule
\multirow{4}{*}{\begin{tabular}[c]{@{}l@{}}DeepSeek-\\ V3\end{tabular}}       & BLUE                    &  0.005  &  0.013  &    0.017   &    0.014      &    0.007  &  0.018    &      0.015   &   0.016        \\
      & ROUGE    &   0.097 &   0.281 &  0.255     &     0.251     &    0.166  &   0.200   &  0.302       &    0.293       \\
      & Token-level F1     &  0.195  &  0.272  &    0.247   &    0.250      &  0.213    &   0.239   &     0.268    &     0.281      \\
      & Cosine   &  0.279  &  0.405  &  0.351     &      0.371    &   0.209   &  0.454    &   0.420      &     0.429      \\
\bottomrule
\end{tabular}
}
\end{table*}

\begin{table*}[!t]
\centering
\caption{Similarity comparison between the \textit{questions section} of academic reviews generated by LLMs, RAG-assisted LLMs, and the \textit{questions section} of human reviews under different text similarity metrics.
In the table, N. is normal, R. is revised, K. is keywords, and S. is sample sentences.
}
\label{table:comments similarity}
\scalebox{0.7}{
\begin{tabular}{c|l|cccc|cccc}
\toprule
\multirow{2}{*}{LLM}                                                             & \multirow{2}{*}{Metric} & \multicolumn{4}{c|}{LLM}   & \multicolumn{4}{c}{LLM with RAGs} \\
&    & N. & R. & R.+K. & R.+K.+S. & N.   & R.   & R.+K.   & R.+K.+S.  \\
\midrule
\multirow{4}{*}{GPT-4o}   & BLUE                    &  0.008  &  0.018  &    0.019   &     0.013     &    0.011  &  0.028    &   0.017      &     0.029      \\
     & ROUGE      &  0.122  & 0.199   &  0.231     &     0.253     &   0.164   &  0.287    &  0.283       &      0.303     \\
      & Token-level F1       &  0.192  &  0.237  &    0.213   &  0.271     &   0.233   &  0.309    &     0.318    & 0.292      \\
      & Cosine             &  0.233  &  0.416  &   0.405    &      0.397    &   0.414   &   0.442   &   0.470      &      0.461     \\
\midrule
\multirow{4}{*}{\begin{tabular}[c]{@{}l@{}}Claude-3.7\\ -Sonnet\end{tabular}} & BLUE                    &  0.009  &  0.006  &    0.014   &   0.021       &   0.012   &  0.018    &  0.024       &   0.047        \\
      & ROUGE     &  0.121  &  0.194  &  0.292     &     0.243     &   0.149   &  0.216    &     0.288    &    0.306       \\
      & Token-level F1     &  0.182  &   0.188 &     0.270  &     0.228     &   0.192   &   0.284   &    0.299     &      0.275     \\
      & Cosine         &  0.310  & 0.393   &  0.382     &    0.374      &    0.318  &   0.499   &      0.442   &     0.415      \\
\midrule
\multirow{4}{*}{\begin{tabular}[c]{@{}l@{}}DeepSeek-\\ V3\end{tabular}}       & BLUE                    &  0.006  &  0.012  &   0.015    &     0.014     &    0.014  &   0.016   &    0.018    &      0.017     \\
      & ROUGE      & 0.094   &  0.238  &   0.295    &    0.239      &   0.120   &   0.219   &  0.287       &    0.259       \\
      & Token-level F1     &  0.225  & 0.196   &  0.302     &    0.289      &  0.151    &  0.288    &    0.261     &    0.284       \\
      & Cosine       &  0.254  &  0.302  &  0.299     &     0.334     &   0.203   &  0.386    &  0.411       &     0.405      \\
\bottomrule
\end{tabular}
}
\end{table*}

\begin{table}[!t]
\centering
\caption{Claim-level correctness evaluation across review conditions.
Lower values indicate better performance.
In the table, N. is normal, R. is revised, K. is keywords, and S. is sample sentences.
}
\label{table:evidence-grounded}
\scalebox{0.7}{
\begin{tabular}{cc|cc}
\toprule
\multicolumn{2}{c|}{\textbf{Methods}} & \textbf{Unsupported Rate} & \textbf{Not-verifiable Rate}  \\
\midrule
\multirow{4}{*}{LLM} & N. & 0.208$\pm$0.032 & 0.311$\pm$0.079 \\
& R. & 0.174$\pm$0.019 & 0.198$\pm$0.041  \\
& R.$+$K. & 0.117$\pm$0.021 & 0.179$\pm$0.016  \\
& R.$+$K.$+$S. & 0.131$\pm$0.026 & 0.163$\pm$0.021 \\
\midrule
\multirow{4}{*}{LLM$+$RAG} & N. & 0.183$\pm$0.017 & 0.351$\pm$0.099 \\
& R. & 0.157$\pm$0.031 & 0.241$\pm$0.111  \\
& R.$+$K. & 0.134$\pm$0.026 & 0.229$\pm$0.051  \\
& R.$+$K.$+$S. & 0.117$\pm$0.036 & 0.191$\pm$0.034 \\
\bottomrule
\end{tabular}
}
\end{table}

\begin{table*}[!t]
\centering
\caption{
Performance comparison of LLMs under different RAG across multiple k values (k=1, 3, 7).
In the table, N. is normal, R. is revised, K. is keywords, and S. is sample sentences.
}
\label{table:ablation}
\scalebox{0.7}{
\begin{tabular}{c|cccc|cccc|cccc}
\toprule
  \multirow{2}{*}{LLM}  & \multicolumn{4}{c|}{k=1} & \multicolumn{4}{c|}{k=3} & \multicolumn{4}{c}{k=7} \\
 & N.     & R.    & R.+K.       & R.+K.+S.    & N.     & R.    & R.+K.       & R.+K.+S.    & N.     & R.    & R.+K.       & R.+K.+S.    \\ \midrule
  & 0.018      & 0.017       & 0.019     & 0.020     & 0.021      & 0.022       & 0.021     & 0.022     & 0.032      & 0.033       & 0.034    & 0.033     \\
 & 0.234      & 0.232       & 0.269     & 0.262     & 0.241      & 0.251       & 0.307     & 0.309     & 0.324      & 0.351       & 0.336    & 0.336     \\
 & 0.219      & 0.230       & 0.241     & 0.259     & 0.232      & 0.290       & 0.259     & 0.266     & 0.309      & 0.333       & 0.317    & 0.330     \\
\multirow{-4}{*}{GPT-4o} & 0.426      & 0.476       & 0.516     & 0.522     & 0.507      & 0.574       & 0.584     & 0.590     & 0.594      & 0.613       & 0.622    & 0.650     \\
\midrule
   & 0.027      & 0.034       & 0.037     & 0.400     & 0.030      & 0.029       & 0.025     & 0.020     & 0.012      & 0.018       & 0.019    & 0.020     \\
   & 0.353      & 0.383       & 0.394     & 0.398     & 0.400      & 0.399       & 0.344     & 0.343     & 0.183      & 0.228       & 0.213    & 0.236     \\
   & 0.322      & 0.347       & 0.401     & 0.403     & 0.346      & 0.344       & 0.335     & 0.286     & 0.241      & 0.259       & 0.261    & 0.241     \\
\multirow{-4}{*}{\begin{tabular}[c]{@{}l@{}}Claude-3.7\\ -Sonnet\end{tabular}}                   & 0.617      & 0.717       & 0.704     & 0.710     & 0.607      & 0.707       & 0.670     & 0.651     & 0.462      & 0.519       & 0.587    & 0.574     \\
\midrule
  & 0.016      & 0.021       & 0.025     & 0.021     & 0.010      & 0.012       & 0.017     & 0.016     & 0.026      & 0.011       & 0.009    & 0.012     \\
 & 0.289      & 0.316       & 0.271     & 0.289     & 0.278      & 0.187       & 0.258     & 0.320     & 0.154      & 0.168       & 0.165    & 0.269     \\
 & 0.273      & 0.264       & 0.254     & 0.240     & 0.150      & 0.330       & 0.231     & 0.244     & 0.128      & 0.279       & 0.250    & 0.233     \\
\multirow{-4}{*}{\begin{tabular}[c]{@{}l@{}}DeepSeek-\\ V3\end{tabular}}                 & 0.525      & 0.540       & 0.571     & 0.583     & 0.441      & 0.617       & 0.647     & 0.690     & 0.384      & 0.535       & 0.525    & 0.557     \\ \bottomrule
\end{tabular}
}
\end{table*}

\mypara{Strengths Section Similarity Analysis}
Similarity scores for the \textit{strengths} section are shown in \autoref{table:strength similarity}.
The revised prompt demonstrates clear effectiveness.
For example, its combination with keywords and sample sentences boosts Claude-3.7-Sonnet to 0.478.
RAG-enhanced technique introduction yields inconsistent responses.
GPT-4o shows significant improvement, particularly with revised prompts, while Claude-3.7-Sonnet experiences dramatic performance decline, and DeepSeek-V3 shows only slight improvement in human similarity.
This pattern reveals specific interactions between RAG and LLMs' internal mechanisms for representing paper strengths.
GPT-4o effectively utilizes retrieved information, while Claude-3.7-Sonnet's retrieval may interfere with its \textit{strengths} construction.
This finding cautions that retrieval enhancement may not benefit all LLMs and tasks equally.

\mypara{Weaknesses Section Similarity Analysis}
In evaluating paper \textit{weaknesses} is shown in \autoref{table:weakness similarity}.
Revised prompts improve \textit{weaknesses}  across all three LLMs, especially Claude-3.7-Sonnet, reaching 0.492 in the revised prompt with the keywords condition in the cosine metric.
Combined RAG and modification techniques worked well for disadvantage descriptions, with Claude-3.7-Sonnet achieving the largest improvement at the cosine metric of 0.429 in the RAG with revised prompts.
The lower similarity scores and smaller differences in disadvantage descriptions indicate common LLM limitations in critical appraisal that RAG techniques only partly resolve.

\mypara{Questions Section Similarity Analysis}
\autoref{table:comments similarity} shows that the \textit{questions} part of the LLM-generated review displays the lowest human similarity, suggesting this is the most challenging aspect of the LLM-generated academic review.
GPT-4o improved most significantly, and Claude-3.7-Sonnet improved slightly with RAG-enhanced and more with additional modifications.
With RAG-enhanced, DeepSeek-V3 performs worse, from 0.254 to 0.203, but improves when using a revised prompt with keywords.
This complex pattern suggests models handle retrieved information differently when writing detailed academic reviews.
GPT-4o integrates retrieved information more efficiently to generate near-human reviews, while DeepSeek-V3 struggles with the RAG-enhanced.
This is because the \textit{questions} part requires more advanced reasoning capabilities and domain knowledge integration, precisely where current LLMs lag behind human thinking.

\begin{table*}[!t]
\centering
\caption{
Impact of RAG and non-RAG on text complexity metrics across different LLMs.
In the table, N. is normal, R. is revised, K. is keywords, and S. is sample sentences.
}
\label{table:linguistic metric}
\scalebox{0.7}{
\begin{tabular}{c|c|cc|cc|cc}
\toprule
\multirow{2}{*}{Metrics} & \multirow{2}{*}{Prompt Types} &  \multicolumn{2}{c|}{GPT-4o} & \multicolumn{2}{c|}{Claude-3.7-Sonnet} & \multicolumn{2}{c}{DeepSeek-V3} \\
 &  &   No Rag  & With RAGs & No Rag  & With RAGs & No Rag  & With RAGs \\
\midrule
\multirow{4}{*}{MTLD}
& N. & 185.631 & 147.380 & 137.664 & 152.331 & 152.506 & 108.804 \\
& R. & 196.319 & 239.379 & 185.149 & 191.779 & 228.502 & 227.430 \\
& R. + K. & 242.777 & 200.162 & 205.529 & 205.152 & 240.575 & 263.645 \\
& R. + K. + S. & 224.301 & 244.630 & 221.161 & 168.163 & 161.087 & 282.309 \\
\midrule
\multirow{4}{*}{Syntactic Complexity}
& N. & 3.638 & 2.398 & 2.984 & 2.761 & 4.127 & 2.742 \\
& R. & 3.182 & 2.769 & 2.622 & 2.942 & 3.557 & 2.531 \\
& R. + K. & 2.541 & 2.630 & 2.143 & 2.657 & 3.037 & 3.011 \\
& R. + K. + S. & 2.863 & 2.726 & 2.421 & 2.307 & 2.842 & 2.135 \\
\bottomrule
\end{tabular}
}
\end{table*}

\section{Ablation Study for Evidence-Grounded Correctness Analysis}
\label{Evidence-Grounded}

We conduct an evidence-tracing experiment to directly evaluate factual faithfulness.
We sample 40 papers and analyze eight review variants per paper.
For each review, we extract 10 claims and classify each as supported, unsupported, or not verifiable using a fixed retrieval and labeling protocol applied uniformly across all conditions~\cite{TBXLKBLMQL25,KMXS20}.

\autoref{table:evidence-grounded} shows that revised prompts consistently reduce unsupported and not-verifiable claims relative to normal prompts across both LLM and RAG settings.
For example, in LLM settings, the unsupported rate decreases from 0.209 to 0.174 and the not verifiable rate from 0.311 to 0.198, with further reductions to around 0.117 and 0.179 when using keyword and sample augmentations.
Similar trends are observed under RAG, although the improvements are less stable.
Keyword and sample augmentations further improve faithfulness, particularly in reducing unsupported claims.

\section{Ablation Study for RAG Paradox}
\label{ablation study}

In RAG, k represents the number of retrieved documents provided as context to the model~\cite{CLHS24}.
We use k=5 as the main setting, while k=1, 3, and 7 serve as ablation studies to understand how retrieval volume affects performance across different prompt types.
We randomly selected 40 papers from our experimental dataset for this ablation study.
For each paper, we collected different LLM-generated reviews to compare.

The results from \autoref{table:ablation} reveal that GPT-4o demonstrates consistent improvement as k increases.
Across all four rows of metrics, its scores generally rise from k=1 to k=7.
The fourth metric shows the most pronounced gains, increasing from 0.426 at k=1 to 0.594 at k=7 under normal prompts.
This suggests GPT-4o effectively leverages additional retrieved context to enhance review quality.
Claude-3.7-Sonnet shows the opposite trend, with performance peaking at lower k values and declining substantially as retrieval volume increases.
Under normal prompts, the second metric drops from 0.353 at k=1 to 0.166 at k=5 and 0.183 at k=7.
This systematic decline indicates that Claude-3.7-Sonnet struggles with processing larger amounts of retrieved information.
DeepSeek-V3 exhibits unstable and generally declining performance, with scores fluctuating inconsistently across k values.
The second metric under normal prompts illustrates this instability: 0.289 at k=1, 0.278 at k=3, 0.148 at k=5, and 0.154 at k=7, showing erratic behavior rather than systematic trends.
These trends confirm that RAG affects models in systematically different ways and reproduces the paradox across retrieval settings.
This analysis provides insight into the conditions under which RAG helps or hinders review quality.

\section{More Details for Linguistic Metric}
\label{linguistic}

We have supplemented our analysis with a linguistic metric to further validate these findings.
We used two metrics: lexical diversity (MTLD)~\cite{MJ10} and syntactic complexity~\cite{L10}.
MTLD quantifies vocabulary richness, with higher scores indicating more diverse word usage.
Syntactic complexity measures sentence structure sophistication, with higher values reflecting more complex grammatical patterns.
Based on \autoref{table:linguistic metric}, the results demonstrate that revised prompts consistently enhance both metrics across all three models.
Comparing prompt types without RAG, the revised (R.), keywords (R.+K.), and sample (R.+K.+S.) versions substantially outperform the normal (N.) baseline in MTLD scores.
For instance, GPT-4o's MTLD increases from 185.631 to 242.777, while Claude-3/7-Sonnet improves from 137.664 to 221.161, and DeepSeek-V3 from 152.506 to 240.575.
This pattern holds universally across models, confirming that structured prompts with keywords and examples elevate output quality beyond simple information retrieval.
These improvements cannot be attributed to stylistic imitation alone.
Syntactic complexity gains are modest and sometimes decline with additional prompt elements, while lexical diversity increases substantially.
This indicates that the revised prompts offer conceptual scaffolding that directs models towards expert-level analysis via profound understanding rather than simple replication.
Different models react differently to the same prompts, which shows that they are actually processing information instead of simply copying it.

\section{More Details for Human Evaluation }
\label{Human evaluation}

\subsection{Human Evaluation Methodology}
\label{HE method}

We randomly selected 40 papers from our experimental dataset for human evaluation.
For each paper, we collected four reviews: one human-written review and three LLM-generated reviews (normal prompt, revised prompt, and RAG-enhanced).
This resulted in 160 total reviews for evaluation.
Three expert annotators independently evaluated all reviews using a blind evaluation protocol, where annotators were unaware of the review source (human vs.\ LLM method).
Each review was assessed using a structured rubric focusing on three key dimensions: insightfulness, constructiveness, and helpfulness~\cite{GSCOABS25,SGSCHB19}.
The evaluation criteria included:
\begin{itemize}
    \item Insightfulness: Depth of analysis and novel observations about the paper.
    \item Constructiveness: Quality and actionability of feedback provided to authors.
    \item Helpfulness: Overall utility for improving the manuscript.
\end{itemize}
We evaluated reviews across five dimensions using a 1-10 scale~\cite{GSMG22}:
\begin{itemize}
    \item 0, Poor quality, Vague, generic comments with no specific insights.
    \item 3, Below average, Identifies some issues but lacks depth and solutions.
    \item 5 Average, Provides moderate detail but limited constructive guidance.
    \item 7, Above average, Offers specific analysis with helpful suggestions.
    \item 10, High quality, Delivers detailed, actionable insights that clearly benefit authors.
\end{itemize}

\begin{table}[!t]
\centering
\caption{Multidimensional human evaluation comparing human-written, normal, revised, and revised-with-RAG reviews across constructiveness, specificity, and faithfulness.
}
\label{table:human multidimensional}
\scalebox{0.7}{
\begin{tabular}{c|cccc}
\toprule
Methods & Most Constructive & Most Specific & Most Faithful \\
\midrule
Human & 47.851\%	& 31.252\%	& 42.504\% \\
Normal & 11.248\%	& 11.633\%	&15.151\% \\
Revised & 28.510\%	& 45.209\%	& 27.463\% \\
Revised $+$ RAG & 12.390\%	& 11.909\%	& 14.885\% \\
\bottomrule
\end{tabular}
}
\end{table}

\subsection{Expanded Human Evaluation With Multidimensional Criteria}
\label{Expanded Human Evaluation}

To complement the main human evaluation, we further conduct a multidimensional comparative study on the same 40 randomly selected papers from the 480-paper test set.
For each paper, we compare four reviews, including human-written, LLM-normal, revised, and revised-with-RAG, resulting in 160 reviews in total.
Annotators with prior peer-review experience are asked to identify the most constructive~\cite{RBG99}, most specific~\cite{SBB25}, and most faithful~\cite{BBCFFHKMPPRRSSWWZ23} review for each paper and to report their confidence level for each judgment.

Human-written reviews are most frequently selected as most faithful (42.50\%), while revised reviews are most frequently selected as most specific (45.21\%) and are also chosen substantially more often as most constructive (28.51\%) than the normal (11.25\%).
Revised-with-RAG performs worse than Revised across all three dimensions, which is consistent with the overall human evaluation results in \autoref{table:human multidimensional}, suggesting relatively stable comparative preferences.

\subsection{Detailed Results}
\label{HE results}

According to \autoref{table:human evaluation}, human-written reviews outperform LLM-generated reviews, averaging 7.12, whereas LLM-generated reviews range from 5.37 to 6.68.
Human-written reviews have 66.95\% high-quality responses, compared to 60.33\% for revised LLM-generated reviews.
This reduces to 42.50\% for the RAG-enhanced LLM-generated review and 31.67\% for the normal LLM-generated version.
Notably, RAG-enhanced LLM reviews exhibit poor performance in high-quality response rates of 42.50\%, suggesting RAG-enhanced techniques may not have achieved the expected enhancement for this job.
For reliability assessment, we calculate agreement as the percentage of samples where all three raters assign identical categories, such as rate as 1, rate as 3/5/7, and rate as 10.
The Fleiss' $\kappa$ is computed to measure inter-rater reliability among multiple raters using the standard formula that accounts for chance agreement~\cite{FQ15}.
The annotation task demonstrates high consistency among three independent raters, with substantial agreement ($\kappa$ = 0.707, 85.00\%~\cite{M12}) across 160 samples.

\begin{figure*}[!t]
\centering
\includegraphics[width=0.90\linewidth]{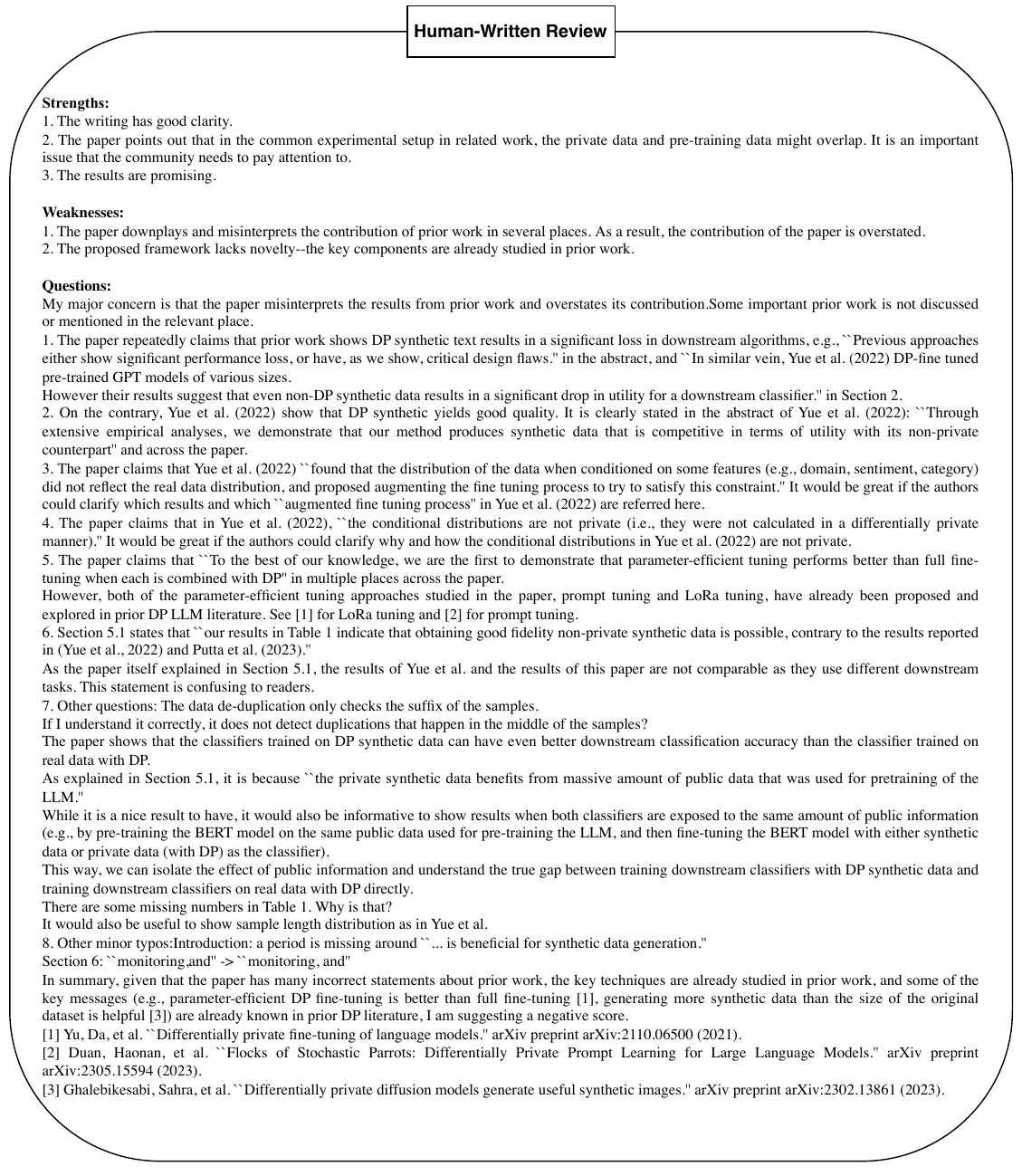}
\caption{Human-Written review.
}
\label{figure:human}
\end{figure*}

\begin{figure*}[!t]
\centering
\includegraphics[width=0.90\linewidth]{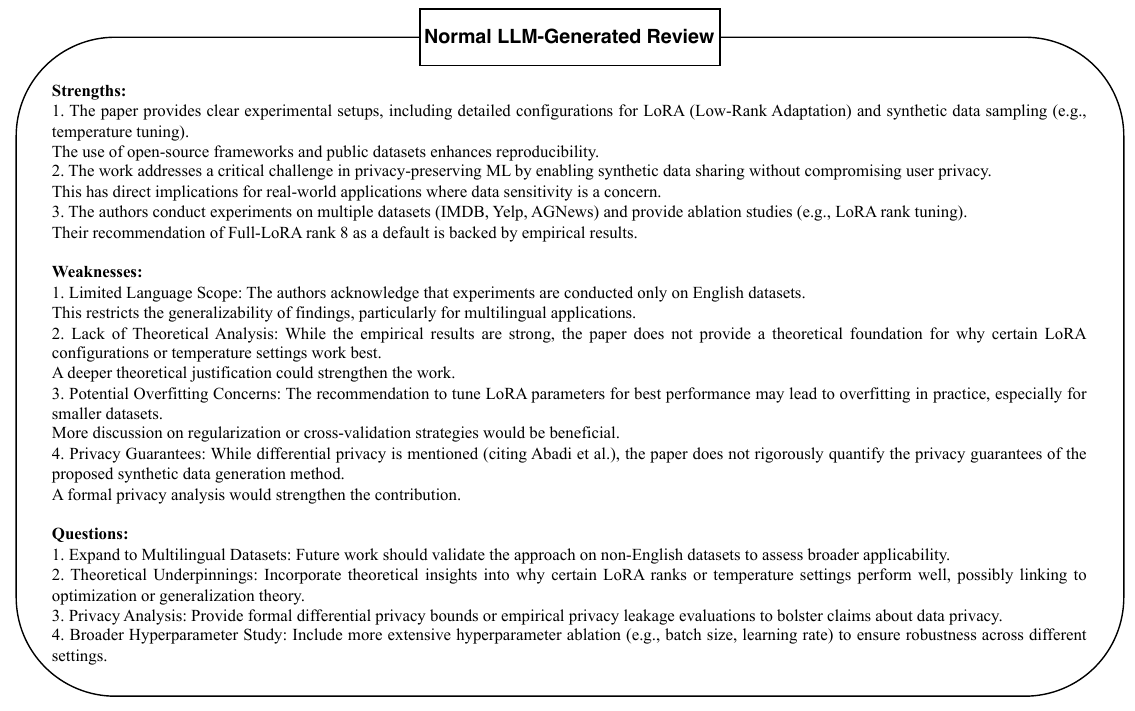}
\caption{Normal LLM-Generated review.
}
\label{figure:normal}
\end{figure*}

\begin{figure*}[!t]
\centering
\includegraphics[width=0.90\linewidth]{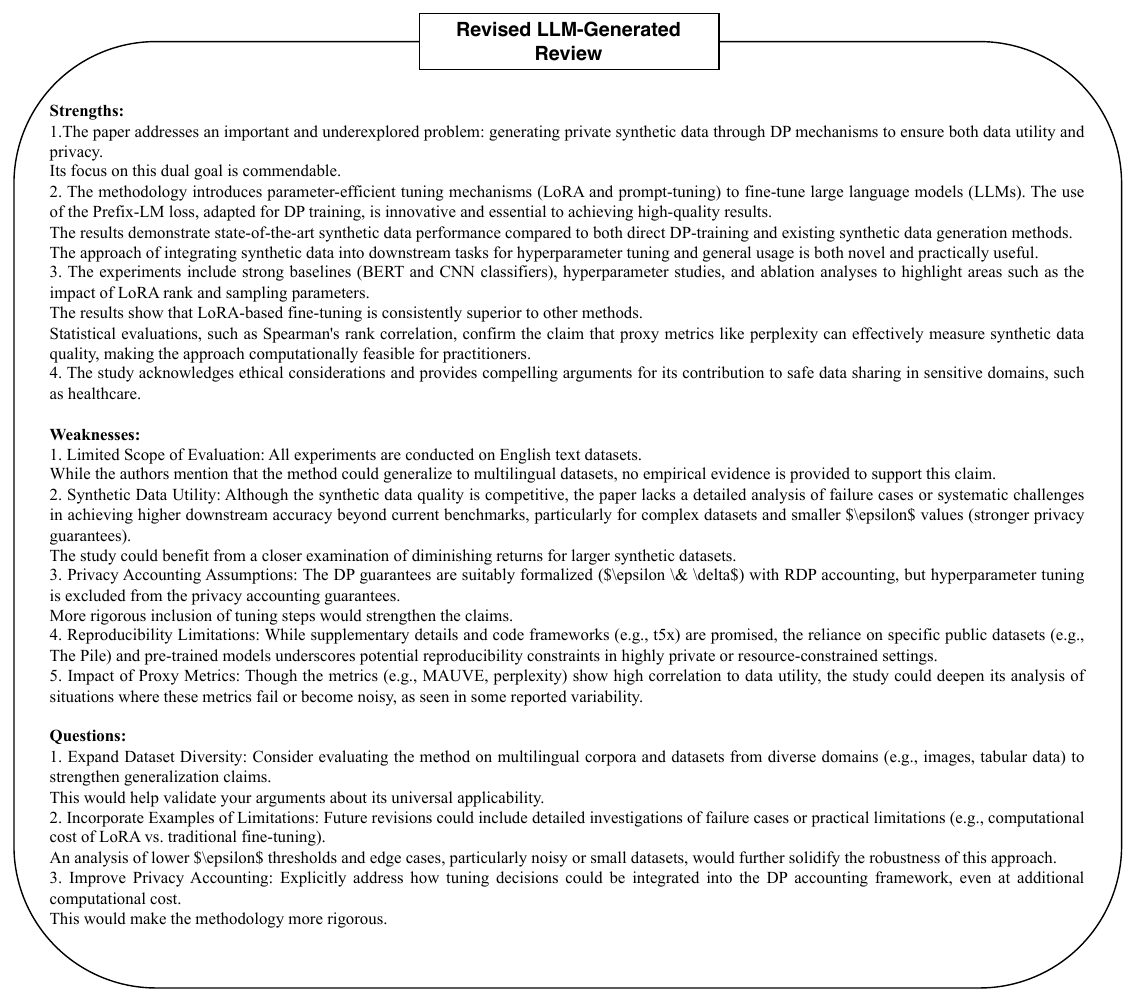}
\caption{Revised LLM-Generated review.
}
\label{figure:revised}
\end{figure*}

\begin{figure*}[!t]
\centering
\includegraphics[width=0.90\linewidth]{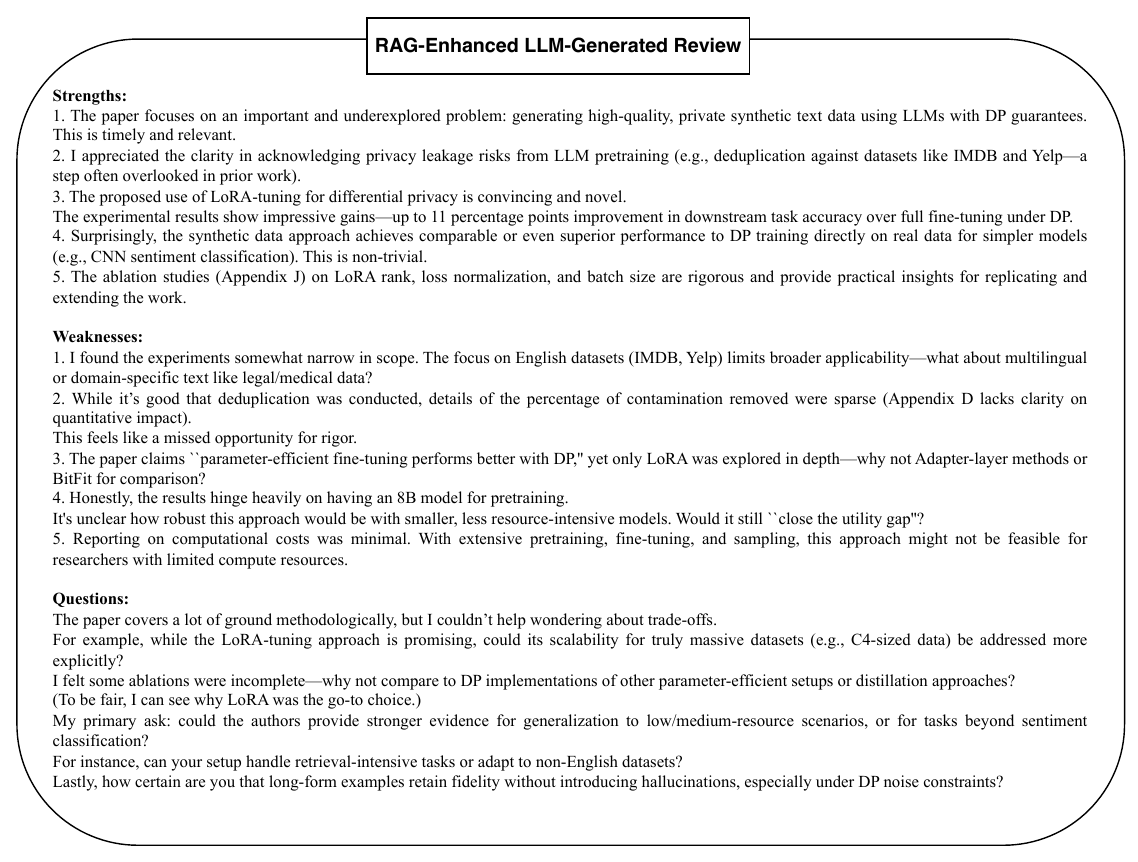}
\caption{RAG-Enhanced LLM-generated review.
}
\label{figure:rag}
\end{figure*}

\subsection{Case Study}
\label{HE case study}

We present a representative example comparing all four review generation methods on the same paper to illustrate quality differences.

The four peer reviews on the same paper demonstrate varied concerns and viewpoints, with each review emphasizing different topics and aspects, as shown in \autoref{figure:human}, \autoref{figure:normal}, \autoref{figure:revised} and \autoref{figure:rag}.

\mypara{Analysis}
In human-written reviews, it is quite strict on academic rigor and criticizes manuscripts on factual accuracy, citation completeness, and originality, especially misunderstandings of past work and overclaims.
The style is nitpicky and questioning-driven, scrutinizing everything from experiments to formatting errors.
The main focus is on verifying if papers accurately describe prior work and whether contributions are truly novel.

The Normal LLM-generated review follows a standardized, procedural evaluation method that looks at how clear, reproducible, and useful the experiments are.
It acknowledges the usage of open-source tools and experimental transparency but suggests that multilingual dataset evaluation could be expanded and privacy analysis should be added.
In general, it is easy on authors' self-statements, not very critical of new ideas and technical specifics, and it gives suggestions that are quite neutral and general.

The revised LLM-generated review is more detailed and critical.
It recognizes the research direction of balancing privacy and utility together with empirical results.
However, it points out some major shortcomings: limited experimental scope, insufficient failure case analysis, and incomplete privacy accounting.
The revised LLM-generated review calls for authors to further demonstrate practical usability, reproducibility, and robustness of evaluation metrics.

The RAG-enhanced LLM-generated review focuses more on practical and implementation details.
It deeply explores method limitations such as computational requirements and incomplete ablation studies.
It also raises forward-looking questions about generalizability to multilingual, non-textual, and low-resource scenarios while emphasizing experimental detail transparency.
This review is defined by its emphasis on practical application, scalability, and methodological rigor.

Overall, all three LLM-generated reviews are constructive and share consensus in recognizing the paper's importance, but differ in focus and depth of criticism.
The normal LLM-generated review is more generic and process-oriented, offering standardized suggestions.
The revised LLM-generated review is more detailed and critical, emphasizing the coexistence of innovation and methodological gaps.
The RAG-enhanced LLM-generated review focuses most on practical applications, resource requirements, and method generalizability.
Although three perspectives offer useful feedback to authors, they still fall short of human reviews in terms of academic rigor and in-depth scrutiny of literature attribution.

\end{document}